\documentclass[runningheads]{llncs}
\usepackage{eccv}
\usepackage{eccvabbrv}

\usepackage{graphicx}
\usepackage{booktabs}
\usepackage{makecell} 
\usepackage{enumitem}
\usepackage{multirow}
\usepackage{xcolor}         
\usepackage{colortbl}
\usepackage{ulem}
\usepackage{wrapfig}

\definecolor{cgray}{RGB}{240,240,240}
\definecolor{mygray}{gray}{.9}

\newcommand{\ours}{GiT\xspace}
\usepackage[accsupp]{axessibility}  

\usepackage{hyperref}

\usepackage{orcidlink}
\usepackage{bbding}
\newcommand{\myparagraph}[1]{\vspace{-1pt}\noindent{\bf #1}}

\begin{document}

\title{\ours: Towards Generalist Vision Transformer through Universal Language Interface}

\titlerunning{GiT: Generalist Vision Transformer}
\author{Haiyang Wang\inst{1,2}$^*$ \and Hao Tang\inst{1}$^*$ \and Li Jiang\inst{3,2}\Envelope \and Shaoshuai Shi\inst{2} \and \\ Muhammad Ferjad Naeem\inst{4} \and Hongsheng Li\inst{5} \and Bernt Schiele\inst{2} \and Liwei Wang\inst{1}\Envelope}

\authorrunning{GiT: Generalist Vision Transformer}
\institute{$^1$Peking University ~~~$^2$Max Planck Institute for Informatics \\ $^3$ The Chinese University of Hong Kong, Shenzhen \\
$^4$ ETH Zurich ~~~$^5$ The Chinese University of Hong Kong\\
\email{\{wanghaiyang@stu, tanghao@stu, wanglw@cis\}.pku.edu.cn}\\ 
\email{jiangli@cuhk.edu.cn ~~\{sshi, schiele\}@mpi-inf.mpg.de} \\
\email{ferjad.naeem@vision.ee.ethz.ch ~~~hsli@ee.cuhk.edu.hk}
}

\def\customsymbol#1{
    \ifcase\number\value{#1}
        \or*
        \or\Envelope
    \else\@ctrerr
    \fi
}

\maketitle

\renewcommand{\footnotesize}{\fontsize{8pt}{8pt}\selectfont}
\renewcommand{\thefootnote}{\customsymbol{footnote}}
\footnotetext[1]{Equal contribution. ~~\Envelope ~~Corresponding author.}
\begin{figure}[h]
    \centering
    \vspace{-20pt}
    \includegraphics[width=0.90\linewidth]{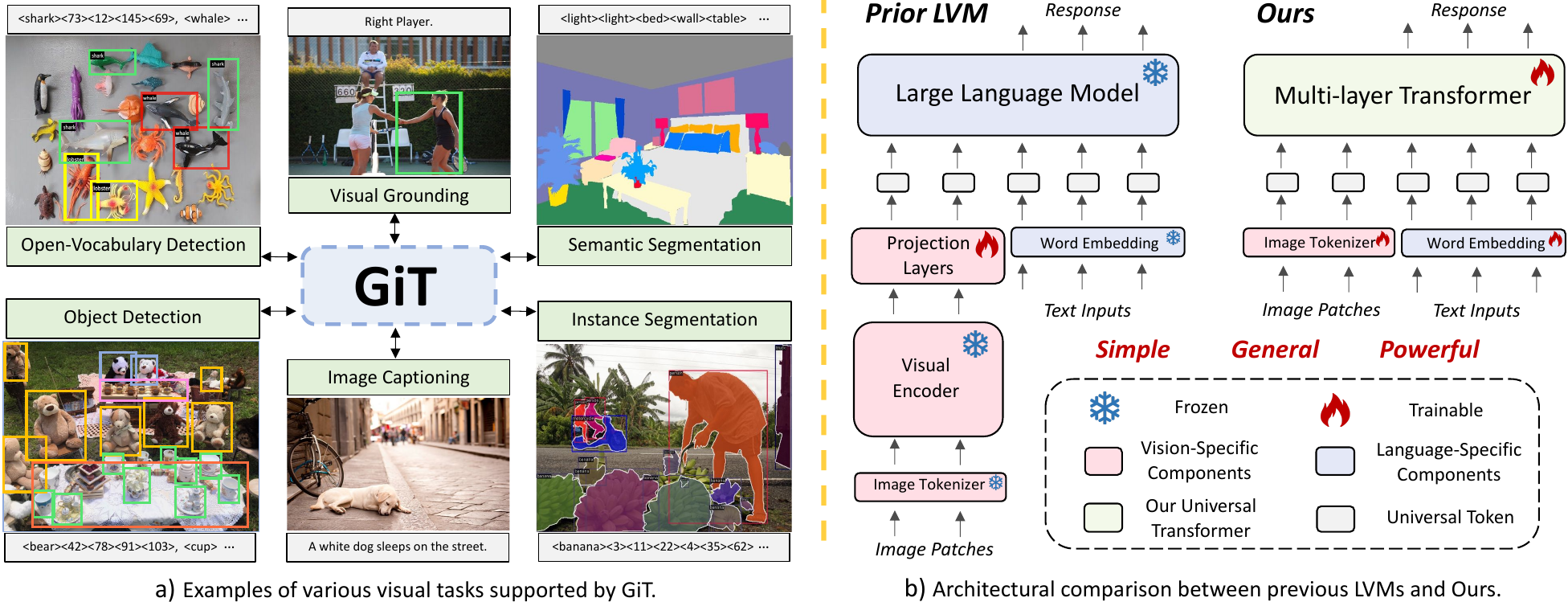}
    \vspace{-8pt}
    \caption{\textit{\uline{\textbf{\normalsize{G}}}eneralist V\uline{\textbf{\normalsize{i}}}sion \textbf{\uline{\normalsize{T}}}ransformer}. a) Examples of tasks supported by \ours. b) Architectural comparison between previous LVMs~(\textit{e.g.}, LLaVA~\cite{liu2023visual}), and ours. \ours seamlessly handles various vision-centric tasks, particularly fine-grained visual perception, via a universal language interface using a plain transformer~(\textit{e.g.}, ViT and GPT).}
    \vspace{-26pt}
    \label{fig_demo}
\end{figure}

\begin{abstract}
This paper proposes a simple, yet effective framework, called \ours, simultaneously applicable for various vision tasks only with a vanilla ViT. Motivated by the universality of the Multi-layer Transformer architecture~(\textit{e.g.}, GPT) widely used in large language models~(LLMs), we seek to broaden its scope to serve as a powerful vision foundation model~(VFM). However, unlike language modeling, visual tasks typically require specific modules, such as bounding box heads for detection and pixel decoders for segmentation, greatly hindering the application of powerful multi-layer transformers in the vision domain.
To solve this, we design a universal language interface that empowers the successful auto-regressive decoding to adeptly unify various visual tasks, from image-level understanding~(\textit{e.g.} captioning), over sparse perception~(\textit{e.g.} detection), to dense prediction~(\textit{e.g.} segmentation). Based on the above designs, the entire model is composed solely of a ViT, without any specific additions, offering a remarkable architectural simplification. \ours is a multi-task visual model, jointly trained across five representative benchmarks without task-specific fine-tuning. Interestingly, our \ours builds a new benchmark in generalist performance, and fosters mutual enhancement across tasks, leading to significant improvements compared to isolated training. This reflects a similar impact observed in LLMs. Further enriching training with 27 datasets, \ours achieves strong zero-shot results over various tasks. Due to its simple design, this paradigm holds promise for narrowing the architectural gap between vision and language. Code and models will be available at \url{https://github.com/Haiyang-W/GiT}.
\end{abstract}

\vspace{-18pt}
\section{Introduction}\label{sec:intro}
\vspace{-4pt}
Developing a universal model capable of completing various tasks aligned with human intention is a long standing goal in Machine Learning. In language processing, the emergence of LLMs~\cite{radford2018improving,touvron2023llama,chowdhery2023palm,zhang2022opt} opens up a promising route, which only employs several stacked transformer layers for adaptable task management with minimal prompts. In this paper, we explore this simple multi-layer transformer~\cite{vaswani2017attention} architecture in visual modeling, seamlessly integrating numerous vision tasks with a universal language interface, aiming to close the architecture gap between vision and language.

The Machine Learning community is undergoing a paradigm shift with the rise of foundation models (\textit{e.g.}, GPT~\cite{brown2020language}, BERT~\cite{devlin2018bert}, DALL-E~\cite{ramesh2021zero}) trained on massive data, enabling the sharing of conceptual knowledge, and offering seamless adaptability to diverse downstream tasks. Language models~\cite{brown2020language,touvron2023llama,devlin2018bert} have greatly benefited from this recently, thanks to a homogenized representation (\textit{i.e.}, input and output are uniformly represented as token sequence). State-of-the-art models like GPT4~\cite{openAI2023gpt4}, LLaMA~\cite{touvron2023llama}, PaLM2~\cite{chowdhery2023palm} and Gemini~\cite{team2023gemini} have shown an unprecedented ability to follow human instructions and solve open-ended tasks. Thanks to their success, this architecture is potentially viewed~\cite{bommasani2021opportunities,reed2022generalist} as a general framework for other machine learning tasks beyond NLP. 

Motivated by this opportunity, the community has developed several large vision models, such as LLaVA~\cite{liu2023visual}, Unified-IO~\cite{lu2023unifiedio} and OFA~\cite{wang2022ofa}, by leveraging vision features~\cite{dosovitskiy2021an,he2016deep} as foreign language of open-source LLMs~\cite{touvron2023llama,alpaca,raffel2020t5}. However, this progress still retained task-specific designs, including diverse visual encoders~\cite{wang2022ofa,zhu2023minigpt}, perception heads~\cite{li2023uni}, RPN~\cite{li2023uni}, and specific target representations~\cite{lu2023unifiedio}.
Task-specific modules require intricate designs for each task a model needs to solve, potentially hindering progress towards a general vision model. Moreover, these task-specific designs typically involve numerous separate training stages~\cite{wang2023visionllm}, complicating model scaling across different tasks. We argue that an alternative general-purpose framework could employ lightweight components through a more universal input-output interface, and allocate most of the model resources to learning a general model across these tasks.

Previous attempts~\cite{zhu2023minigpt,liu2023visual,instructblip,alayrac2022flamingo,li2022blip,wang2022image,fuyu-8b} on large visual modeling predominantly focused on the image-level vision-language domain, mainly due to its straightforward integration into LLMs by viewing the image as a foreign language. This approach often overlooks the incorporation of classical perception tasks, such as detection and segmentation. Developing a unified framework for fundamental visual perception has proven to be quite challenging since it requires the model to predict multiple outputs with different formats in parallel, with annotations varying widely in representations, ranging from coarse-grained image level to fine-grained pixel level. For example, detection yields variable numbers of bounding boxes, segmentation produces binary masks, and image captioning generates textual answers. These drawbacks make it difficult to design a single model simultaneously applicable across all visual tasks. 
\begin{table*}[t]
  \vspace{-4pt}
  \caption{
  Columns from left to right display task source examples, dataset counts, total samples, percentages, and multi-task sampling rates, then task modalities. Highlighted rows summarize statistics for similar task groups. See appendix for the complete list.}
  \label{tab:all_datasets}
  \vspace{-8pt}
  \centering
  \resizebox{0.9\textwidth}{!}{
  \begin{tabular}{l|c|cccc|cc|ccc}
    \toprule
    & Example & \multicolumn{4}{c|}{Size} & \multicolumn{2}{c|}{Input Modalities} & \multicolumn{3}{c}{Output Modalities} \\
    &Sources & Dataset & Size & Percent & Weight & Text & Image & Text & Sparse & Dense \\
    \midrule
    \rowcolor{cgray} \textbf{Image-Level} &  & \textbf{10} & \textbf{11.4m} & \textbf{67.1} & \textbf{40} &\checkmark &\checkmark & \checkmark & \checkmark & - \\
    ~~Image Captioning & \textit{CC12M~\cite{changpinyo2021conceptual},~VG~\cite{krishna2017visual},~SBU~\cite{ordonez2011im2text}}   & 5   &11.3m & 66.6 & 30 & - & \checkmark & \checkmark & - & - \\
    ~~Visual Grounding & \textit{RefCOCO~\cite{yu2016modeling},~Flickr30k~\cite{plummer2015flickr30k}} & 5   &115k  & 0.7 & 10  & \checkmark  & \checkmark & - & \checkmark & - \\
    \rowcolor{cgray}\textbf{Object-Level} &  & \textbf{11} & \textbf{5.2m} & \textbf{30.9} & \textbf{40} & - &\checkmark & - & \checkmark & \checkmark \\
    ~~Object Detection          & \textit{Objects365~\cite{shao2019objects365},~COCO~\cite{lin2014coco}} &   8  & 3.8m & 22.6 & 20 & - &\checkmark & - & \checkmark & - \\
    ~~Instance Segmentation     & \textit{OpenImages~\cite{kuznetsova2020open},~LVIS~\cite{gupta2019lvis}} &   4  & 1.4m & 7.9 & 20 & - &\checkmark & - & \checkmark & \checkmark \\
    \rowcolor{cgray}\textbf{Pixel-Level} &  & \textbf{6} & \textbf{322k} & \textbf{2.0} & \textbf{20} & - &\checkmark & - & - & \checkmark \\
    ~~Semantic Segmentation      & \textit{COCOStuff~\cite{caesar2018coco},~ADE20K~\cite{zhou2017scene}} & 6   & 322k & 2.0 & 20 & - &\checkmark & - & - & \checkmark \\
    \rowcolor{cgray}\textbf{All Tasks} &    & \textbf{27} & \textbf{17m} & \textbf{100} & \textbf{100} & \checkmark &\checkmark & \checkmark & \checkmark & \checkmark \\
    \toprule
  \end{tabular}
  }
  \vspace{-16pt}
\end{table*}

Recent developments in LLMs~\cite{brown2020language,openAI2023chatgpt,openAI2023gpt4,radford2019language} have shown the potential of Transformer~\cite{vaswani2017attention} being a universal computation architecture. Inspired by this, we introduce \ours, a vision foundation model that can handle diverse vision-centric tasks.
As shown in Figure \ref{fig_demo}, compared to previous unified models~\cite{wang2023visionllm,lu2023unifiedio,wang2022ofa}, our method features a minimalist design, comprising just several Transformer layers without any vision-specific additions other than the patch projection layers, closely aligning with LLM architectures. Similar to language modeling, all visual tasks are structured into an auto-regressive framework through a universal language interface. Specifically, our targets are expressed as token sequences using a unified representation, relying solely on a standard vocabulary without involving extra tokens~\cite{wang2023visionllm,reed2022generalist}. To be compatible with various visual tasks across different perceptual scales, we introduce a flexible multi-task template for task unification. It partitions the whole image into $N$ subregions by grid sampling and concurrently processes each subregion with efficient parallel decoding.

The above designs facilitate multi-task training of our model across five representative benchmarks without task-specific fine-tuning. As shown in Table \ref{tab:analysis_multitask} and \ref{tab:all_results}, leveraging shared parameters and representation, our model achieves strong generalist results and mirrors the multi-task capabilities of LLMs~\cite{radford2019language}. Tasks with overlapping abilities can mutually enhance each other, leading to significant gains over separate training~(see \S\ref{sec:indistbenckmarking} for more analysis). To further enhance generalizability, we incorporate 27 standard visual datasets into training~(see Table \ref{tab:all_datasets}), resulting in strong zero- and few-shot performances on unseen data.

In particular, our work makes the following contributions:
\begin{itemize}[leftmargin=12pt,itemindent=0pt]
    \vspace{-8pt}
    \setlength{\itemsep}{3pt}
    \setlength{\parsep}{0pt}
    \setlength{\parskip}{0pt}
    \item \textit{Foundational framework for unified visual modeling.} We introduce a simple visual modeling paradigm with a straightforward multi-layer transformer, greatly simplifying the model design. Our model integrates various vision-centric tasks, especially the often-neglected object- and pixel-level tasks, via an efficient universal language interface. 
    \item \textit{Multi-task ability like LLMs.} Weight-sharing and unified learning objectives enable us to obtain the multi-task capability as observed in LLMs, achieving the best and mutually enhanced generalist performance over five benchmarks.
    \item \textit{Strong generalizability.} Fully embracing the one-stage joint training strategy as used in LLMs, our model is trained on 27 publicly available datasets, achieving strong zero- and few-shot performance across various tasks.
    \vspace{-4pt}
\end{itemize}
\begin{table*}[t]
\begin{minipage}[t]{0.52\linewidth}
  \makeatletter\def\@captype{table}
  \caption{Summary of architecture configuration. Shared parameters account for over 98\% of the whole model. The parameter of text embedding is excluded because it operates in a zero-computation index manner.}
  \label{tab:model_param}
  \vspace{-8pt}
  \centering
  \resizebox{1.0\linewidth}{!}{
  \begin{tabular}{l|c|c|c|c|c|c}
    \toprule
     & \multicolumn{3}{c|}{Multi-Modal Tokenizers}  &  Multi-layer & Layer & Total \\
    \multirow{-2}{*}{Model} & Text & Image & Out-of-vocab  &  Transformer & Number & Parameter \\
    \midrule
    \ours$_{\text{Base}}$      &0& 0.4\% & 1.8\%& 97.8\% & 18~(12+6) & 131M\\
    \ours$_{\text{Large}}$      &0& 0.2\% & 1.1\%& 98.7\% & 30~(24+6) & 387M\\
    \ours$_{\text{Huge}}$      &0& $\textless$ 0.1\% & 0.8\% & 99.1\% & 38~(32+6) & 756M\\
    \toprule
  \end{tabular}
  }
\end{minipage}
\hfill
\begin{minipage}[t]{0.48\linewidth}
\makeatletter\def\@captype{table}
  \caption{Abilities required for each task and the performance improvements after multi-task training. $\dag$ means polygon-based segment~\cite{xie2020polarmask,xu2019explicit}, different from the popular mask-based methods~\cite{he2017mask}.}
  \label{tab:analysis_multitask}
  \vspace{-8pt}
  \centering
  \resizebox{1.0\linewidth}{!}{
  \begin{tabular}{l|c|c|c|c|c}
    \toprule
    Task           & Image  & Language & Segment & Localization & \textit{Improve}~\text{(single}$\rightarrow$\text{multi)} \\
    \midrule
    Detection           & \checkmark   & -          &-           &\checkmark  & +\textbf{\textit{1.6}}@AP \\
    InsSeg       & \checkmark   &-           & \color{gray}\checkmark$^\dag$  &\checkmark  & +\textbf{\textit{1.6}}@AP$_{50}$, +\textbf{\textit{0.2}}@AP$_{75}$ \\
    Grounding           & \checkmark   &\checkmark  & -             &\checkmark  & +\textbf{\textit{2.5}}@Acc\\
    Caption      & \checkmark   &\checkmark  & -             & -           & +\textbf{\textit{4.7}}@CIDEr\\
    SemSeg       & \checkmark   &-           & \checkmark  & -           & +\textbf{\textit{0.1}}@mIoU\\
    \toprule
  \end{tabular}
  }
\end{minipage}
\vspace{-16pt}
\end{table*}
\section{Related Work} \label{relatedwork} 
\myparagraph{Multi-layer Transformer}~\cite{vaswani2017attention} has emerged as a universal learning architecture, becoming a cornerstone in most LLM frameworks. Notable LLMs like GPT series~\cite{radford2018improving,radford2019language,brown2020language,openAI2023gpt4,openAI2023chatgpt,ouyang2022instrucGPT}, as well as LLaMA~\cite{touvron2023llama}, PaLM~\cite{chowdhery2023palm}, and OPT~\cite{zhang2022opt} have made significant advances in this domain. Beyond language, plain transformer also has proven effective in 2D vision with ViT~\cite{dosovitskiy2021an}, 3D vision via DSVT~\cite{wang2023dsvt}, multimodal imaging in UniTR~\cite{wang2023unitr}. Despite their success, these straightforward transformers are often limited to feature encoding and require task-specific modules, greatly hindering the progress toward a general learner. To solve this, we aim to broaden the scope of multi-layer transformer, moving beyond their conventional encoder-only function to an LLM-like visual modeling. Our model employs several transformer layers for various visual tasks with a universal language interface, narrowing the architectural gap between the vision and language. 

\myparagraph{Vision Foundation Model} excels in handling diverse visual tasks within a unified architectural framework. Motivated by the success of seq2seq models in NLP, innovations like OFA~\cite{wang2022ofa}, Flamingo~\cite{alayrac2022flamingo}, LLaVA~\cite{liu2023visual} and Gato~\cite{reed2022generalist} have reframed vision-language tasks as sequence generation problems, which is further developed by Unified-IO~\cite{lu2023unifiedio}, Pix2Seq v2~\cite{chen2022unified}, and VisionLLM~\cite{wang2023visionllm} to process spatial information across more tasks. However, these approaches face challenges such as inefficient inference from non-parallel decoding~\cite{chen2022unified} or the complexity of vision-specific additions~\cite{li2023uni,wang2023visionllm,lu2023unifiedio}, slowing progress towards a universal vision model. Moreover, they often lack LLMs' multi-task capabilities, where joint training yields superior performance compared to individual training.
\section{Universal Language Interface} \label{sec:uli}
\begin{figure*}[t]
\vspace{-2pt}
\begin{center}
   \includegraphics[width=0.85\linewidth]{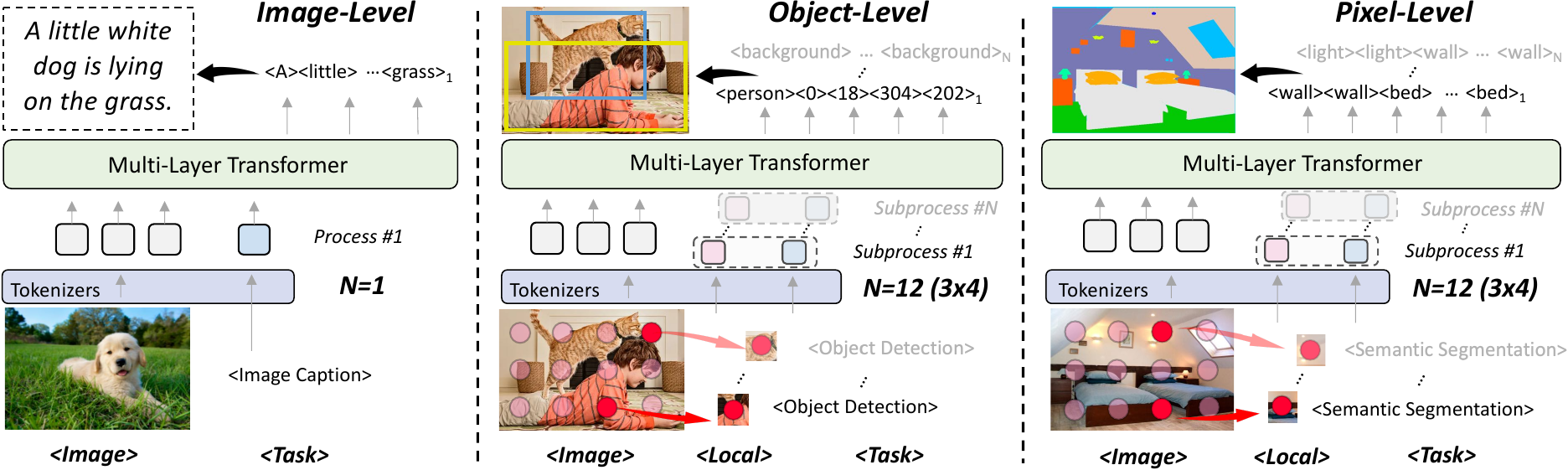}
\end{center}
\vspace{-18pt}
\caption{Task-level customization spans from image- to object- and pixel-level, setting $N$ to 1, 625~(25$\times$25), and 1764~(42$\times$42), in real implementation. Red point means localized visual token, generated by image bilinear interpolation at its grid point. Task prompt is text, converted into a token via text and out-of-vocabulary representation.}
\vspace{-16pt}
\label{fig:task-level-demo}
\end{figure*}

In this section, we propose a simple universal language interface that integrates five fundamental visual tasks, ranging from image, over object to the pixel level, into the successful auto-regressive framework. All our targets are expressed as token sequences via a unified representation~(\S\ref{sec:IO}), and then organized by a general multi-task template~(\S \ref{sec:multi-task template}), which partitions the fine-grained visual perception into a series of parallel-decoded subproblems. 
Figure~\ref{fig:task-level-demo} illustrates the multi-task input templates for three tasks, namely image captioning (image-level task, left), object detection (object-level task, middle) and semantic segmentation (pixel-level task, right). Further technical details are provided below. 
\vspace{-8pt}
\subsection{Unified Input and Output Representation} \label{sec:IO}
To support various modalities such as images, language, bounding boxes, masks, \textit{etc}, it's essential to represent them in a unified space. To achieve this, we straightforwardly project the input image and text into patch and language token embeddings. Following this, all targets are represented via a universal language interface and tokenized entirely based on a standard vocabulary~\cite{wu2016google}.\\
\myparagraph{Text representation.} Vision-language tasks often require text processing, like image captioning, where a natural language description is generated based on the given image. To handle it, we follow the practice of BERT~\cite{devlin2018bert}, texts are transformed into WordPiece~\cite{wu2016google} subwords, with a $\sim$30,000 token vocabulary, and then embedded via a lookup table into a learnable embedding space. Position encodings are added to indicate local positions within time steps.\\
\myparagraph{Out-of-vocabulary representation.} Visual perception typically relies on complex textual concepts comprised of multiple pieces, such as ``\uline{traffic} \uline{light}'' and ``\uline{20}\,\uline{47}'', the category name and numerical value used in object detection. As discussed in \cite{wang2023visionllm,li2022glip}, using multiple tokens to represent them is inefficient. 1) Adding separators like \texttt{$<$/c$>$}``\uline{traffic} \uline{light}''\texttt{$<$/c$>$} to identify categories will extend sequence length, particularly impractical for dense prediction tasks. 2) Varying token length for multi-piece words leads to inconsistent decoding steps, necessitating complex and rule-based post-processing to achieve reliable outcomes. To tackle this problem, some solutions~\cite{wang2023visionllm,reed2022generalist,lu2023unifiedio} introduce new tokens of category and number terms while facing challenges when considering token capacity constraints. Instead of expanding the vocabulary, we treat multi-piece concepts as continuous text and compress them into a single token as follows, 
\begin{equation}
    \footnotesize
    \begin{aligned}
    \mathcal{I}_0, \mathcal{I}_1 & = \text{Tokenizer}(``\text{\uline{traffic}~\uline{light}}"), ~~~~~~~~~~~~~~~~~~~~~ \mathcal{I}~~\text{is the token index,} \\ 
    \mathcal{F}_{0}, \mathcal{F}_{1} & = \text{Attention}(\text{TE}(\mathcal{I}_0)+\text{PE}(\text{0}), \text{TE}(\mathcal{I}_1)+\text{PE}(\text{1})), ~~~~\mathcal{F}_{\text{traffic light}} = \mathcal{F}_{0},\\
    \end{aligned}
\end{equation}
where $\text{Attention}(\cdot)$ is a single-layer attention, $\text{TE}(\cdot)$ and $\text{PE}(\cdot)$ are text and position embedding functions. Our approach offers an alternative solution for handling any out-of-vocabulary terms without expanding the basic vocabulary, which greatly simplifies the post-processing for achieving effective perception. \\
\myparagraph{Sparse representation.} In the context of sparse object-level perceptions such as object detection~\cite{girshick2015fastrcnn,carion2020detr} and instance segmentation~\cite{he2017maskrcnn}, which generate various category and location representations (for example, bounding boxes and instance masks), we propose a standardized output format. This format is defined as a tuple $(C, P)$, where $C$ represents the category label, and $P\text{=}\{x_i, y_i\}^{N}_{i\text{=}1}$ denotes a set of $N$ points that identify the object's location. To align with the format of linguistic tokens, both the class and location targets are tokenized by the prior text and out-of-vocabulary representation. Following VisionLLM~\cite{wang2023visionllm}, continuous coordinates are uniformly discretized into integers within \texttt{[-range, range]}. A bounding box is formulated with four points as $\{x_1, y_1, x_2, y_2\}$, representing its top-left and bottom-right coordinates, while instance mask defines its fine-grained region via multiple points along the boundary~\cite{xie2020polarmask,xu2019explicit}. \\
\myparagraph{Dense representation.} Various perceptual tasks, such as semantic segmentation~\cite{long2015fcn,ronneberger2015unet}, require models to generate dense outputs, often involving per-pixel predictions. To handle these tasks, we start by converting per-pixel labels into unified tokens. For example, semantic classes~\cite{lin2014coco} are firstly tokenized by text and out-of-vocabulary representation. Then, these dense labelings are flattened into a 1D sequence in raster order, represented autoregressively, similar to iGPT~\cite{chen2020igpt}.\\
\myparagraph{Image representation.} Images are converted into a non-overlapping 16 $\times$ 16 patch sequence in raster order and then embedded to tokens with a trainable linear projection and a learnable positional embedding, as done in ViT~\cite{dosovitskiy2021an}. 
\begin{figure*}[t]
\vspace{-2pt}
\begin{center}
   \includegraphics[width=0.8\linewidth]{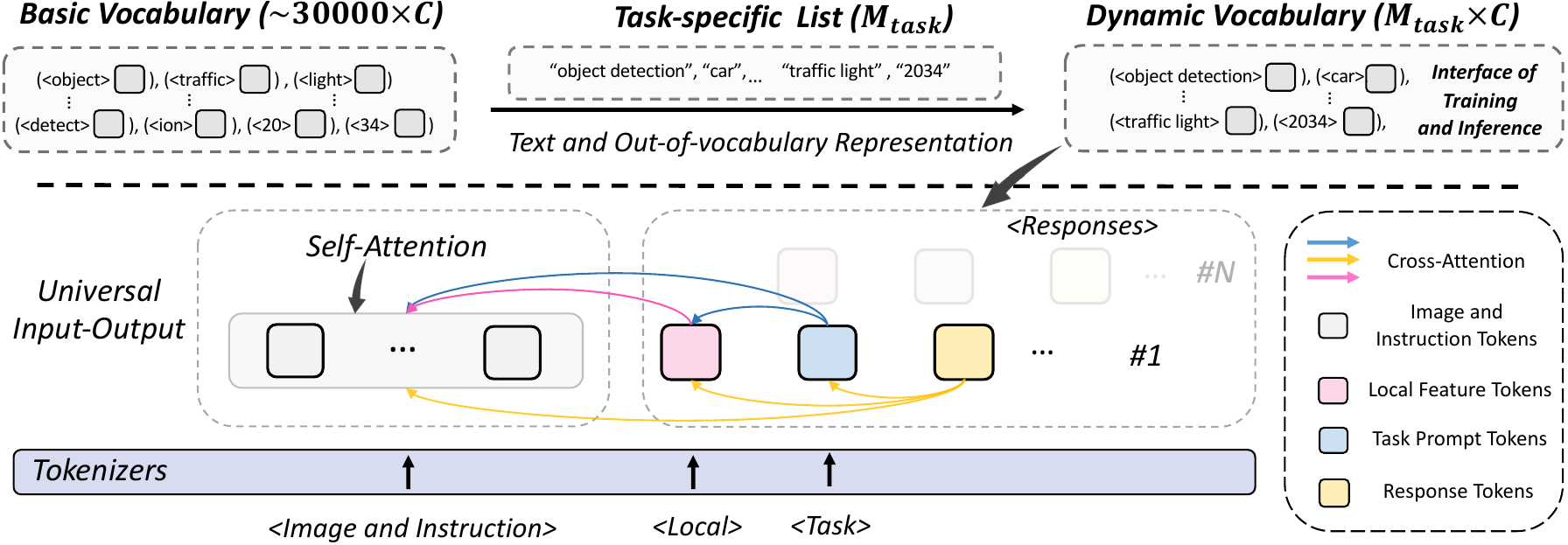}
\end{center}
\vspace{-16pt}
\caption{Our multi-task formulation is broadly illustrated as processing four types of user inputs: image patches, instructive language tokens, and $N$ parallel point-based subprocesses, each with its interpolated local image feature and task identifier for efficient parallel visual prediction. As for the language interface, we use a basic vocabulary, a specific vocabulary list required by the current task, and the task-agnostic out-of-vocabulary module~(\S\ref{sec:IO}) to dynamically create vocabulary sets for each task.}
\vspace{-16pt}
\label{fig_overview}
\end{figure*}
\vspace{-6pt}
\subsection{Multi-Task Template with Parallel Decoding} \label{sec:multi-task template}
Prior to constructing the templates, we first divide 2D visual understanding into three distinct categories, each defined by their perceptual granularity and output representation. Our focus encompasses five core tasks for training and analysis: 1) \textit{Image-level} tasks, exemplified by Image Captioning and Visual Grounding, 2) \textit{Object-Level} tasks like Object Detection and Instance Segmentation, and 3) \textit{Pixel-Level} tasks such as Semantic Segmentation. Then, we introduce a unified seq2seq framework that seamlessly integrates various task formulations, from purely visual to those involving language, enabling flexible task customization. \\
\myparagraph{General Formulation.} Inspired by well-established language models, we adapt the widely accepted instruction template of LLMs to the vision community~(\textit{e.g.}, vision-language and spatial-aware visual perception).  As shown in Figure \ref{fig:task-level-demo} and \ref{fig_overview}, the instructional template is defined as follows,
\begin{equation}
    \footnotesize
    \begin{aligned}
    \underbrace{\texttt{$<$}\text{Image}\texttt{$>$} \texttt{$<$}\text{Instruction}\texttt{$>$}}_{\text{shared global observation}} & 
    \underbrace{
    \begin{cases} 
    \texttt{$<$}\text{LocalFeature$_1$}\;\texttt{$>$} \texttt{$<$}\text{Task$_1$}\;\texttt{$>$} :\texttt{$<$}\text{Response$_1$}\;\texttt{$>$} \\
    \quad \quad \quad  \ \ \vdots \quad \quad \quad \quad \quad \quad  \vdots \quad \quad \quad \quad \quad \ \ \vdots \\
    \texttt{$<$}\text{LocalFeature$_N$}\texttt{$>$} \texttt{$<$}\text{Task$_N$}\texttt{$>$} :\texttt{$<$}\text{Response$_N$}\texttt{$>$}.
    \end{cases}
    }_{\text{multi-track local observations and responses}}
    \end{aligned}
\end{equation}
In our template, user input is structured into four segments. The first comprises image patches, as done in ViT. The second involves instruction inputs, like language expression used for visual grounding. For the third and fourth segments, targeting efficient object- and pixel-level visual perception like simultaneously predicting multiple bounding boxes as in traditional object detection, we partition the task into $N$ parallel local subprocesses by grid sampling, as shown in Figure \ref{fig:task-level-demo}. Each subprocess works with a local image token, created by bilinearly interpolating image features based on its grid point position, and a pure text task identifier, converted into a single token via text and out-of-vocabulary representation. For Vision-Language tasks, we set $N$ to 1, while for vision-centric tasks like detection and segmentation, $N$ is adjustable to match the required prediction resolution. These designs allow our method to flexibly handle nearly all 2D vision tasks. Notably, some segments are optionally required by different tasks, \textit{e.g.}, image captioning only requires image inputs and a task prompt. \\
\indent In contrast to the traditional encoder and decoder setups, we employ various mask matrices to determine the token representation context. As shown in Figure \ref{fig_overview}, our method processes inputs~(\textit{i.e.}, image and instruction) by applying bidirectional self-attention, similar to a typical encoder. Importantly, we enable image-to-text attention to enhance its ability of text-conditioning image processing~(see Table \ref{tab:ablate_experts}). As for computing local and task prompts, and target prediction of each subprocess, we use left-to-right unidirectional attention for modeling causal relations, in line with decoder-only autoregressive approach. \\
\myparagraph{Image-Level.} The definition for image-level tasks such as image captioning and visual grounding is straightforward, closely mirroring the NLP tasks. Following previous vision-language methods, we set $N$ to 1 and structure the token sequence of image captioning as \{\texttt{$<$image$>$} ``\textit{image captioning}''\text{:} \texttt{$<$text$>$}\}, and visual grounding as \{\texttt{$<$image$>$} \texttt{$<$instruction$>$} ``\textit{visual grounding}''\text{:} \texttt{$<$bbox$>$}\}.\\ 
\myparagraph{Object-Level.} 
Developing a generative framework that adeptly manages classical object-level perception tasks, including object detection and instance segmentation, presents a significant challenge. It demands a model capable of concurrently generating all the bounding boxes and masks. To address this, as shown in Figure \ref{fig:task-level-demo}, we introduce a point-based parallel decoding framework designed for visual prompt perception. It starts by sampling a grid of $N$ points across the image, where $N$ is set to 625, corresponding to a 25 $\times$ 25 sampling resolution for 1120 $\times$ 1120 images. Following this, we conduct generative perception at each point using the format: \{\texttt{$<$image$>$} \texttt{$<$local feature$>$} \texttt{$<$task identifier$>$}\text{:} \texttt{$<$sparse response$>$}\}. \texttt{$<$image$>$} is the patch tokens shared by all grid subprocesses. 
\texttt{$<$sparse response$>$} indicates our chosen object-level sparse representation as detailed in \S\ref{sec:IO}. Notably, if the point is in the negative part, \texttt{$<$background$>$} token will be predicted.\\
\begin{figure*}[t]
\vspace{-2pt}
\begin{center}
   \includegraphics[width=0.85\linewidth]{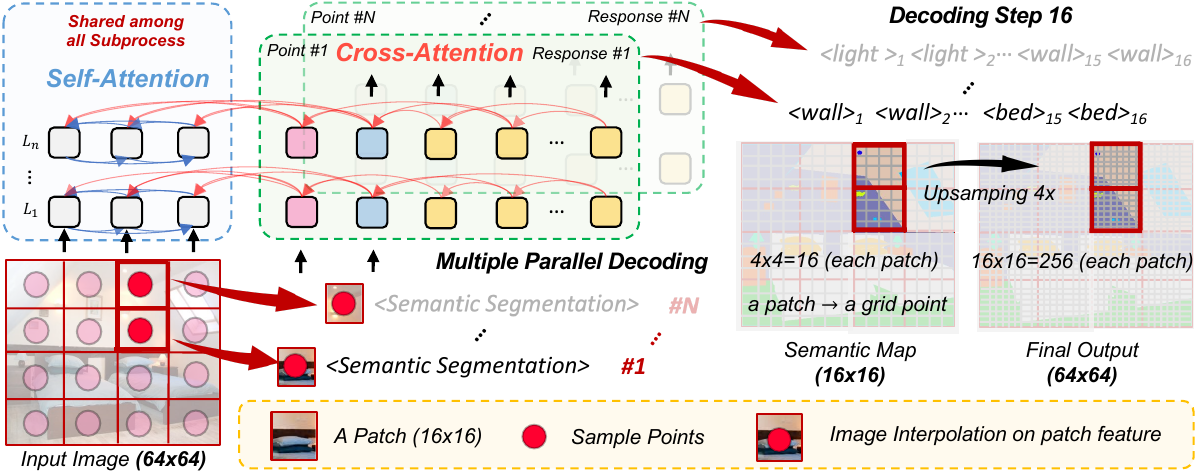}
\end{center}
\vspace{-14pt}
\caption{An illustration of pixel-level multiple parallel decoding. Consider a 64$\times$64 image divided into 16 patches, where each patch is 16$\times$16. With $N$=16 and a decoding step of 16 per subprocess, each grid point covers one patch to predict a 4$\times$4 semantic map, which is then upsampled 4$\times$ to the original size for the final result. }
\vspace{-14pt}
\label{fig_seg_demo}
\end{figure*}
\indent An example of detection for a grid point: \{\texttt{$<$image$>$}\texttt{$<$local feature$>$} ``\textit{object detection}''\text{:} \texttt{$<$c$>$}\texttt{$<$$\text{x}_1$$>$}\texttt{$<$$\text{y}_1$$>$}\texttt{$<$$\text{x}_2$$>$}\texttt{$<$$\text{y}_2$$>$}\}, where \texttt{$<$c$>$} is the class label, and (\texttt{$<$$\text{x}_1$$>$}\texttt{$<$$\text{y}_1$$>$}\texttt{$<$$\text{x}_2$$>$}\texttt{$<$$\text{y}_2$$>$}) indicate the box points' offsets from the grid points. \\
\myparagraph{Pixel-Level.} The auto-regressive decoding paradigm~\cite{openAI2023gpt4,radford2018improving,brown2020language} struggles with high-dimensional outputs, particularly in cases like computing all pixel semantic categories in a single sequence, incurring considerable computational overhead. Earlier efforts~\cite{lu2023unifiedio,ning2023all} attempted to alleviate this using compressed tokens via VQ-VAE~\cite{van2017vqvae}. However, this approach compromised the pure language interface and introduced intricate modules. To tackle this issue, as illustrated in Figure \ref{fig_seg_demo}, we convert per-pixel labels into linguistic tokens and further divide the image into $N$ uniform sub-regions, just like object-level tasks. Specifically, for segmentation tasks, we set $N$ to 1764 to achieve a 42$\times$42 perceptual resolution for images sized 672 $\times$ 672. Each subprocess independently conducts sequential pixel-level predictions in parallel, leading to enhanced efficiency. \\
\indent An example of semantic segmentation for a single track with 16 decoding steps: \{\texttt{$<$image$>$} \texttt{$<$local feature$>$} ``\textit{semantic segmentation}''\text{:} \texttt{$<$c$_1$$>$} \texttt{$<$c$_2$$>$} $\cdot\cdot\cdot$ \texttt{$<$c$_{15}$$>$} \texttt{$<$c$_{16}$$>$}\}, where \texttt{$<$c$_i>$} is the $i\text{-th}$ class token of each sub-region.
\section{Training}
\subsection{Architecture: Multi-layer Transformer} 
By employing the universal language interface, we formulate a diverse array of 2D vision tasks as sequences of discrete input and output tokens. This method has paved the way for extending the successful architectures (such as Multi-layer Transformers~\cite{vaswani2017attention,brown2020language,radford2018improving}) in Large Language Models, to unified visual modeling. \\
\indent Building on the visual foundations, we leverage the structure of window-based ViT~\cite{li2022exploring,dosovitskiy2021an}, identical to the visual encoder used in SAM~\cite{kirillov2023segment}, for both linguistic sequences and high-resolution images. 
A few global attention blocks are evenly integrated into the model for feature propagation. Notably, within the window attention layer, each patch token only interacts with grid points located in the same window. Our approach can be built upon such a common structure~(\textit{i.e.}, ViT) without architectural changes, enhancing the framework's universality. \\
\indent Benefiting from the above designs, our architecture can allocate the most of computational parameters~($\textgreater$ 98\%) to general inference, complemented by a few lightweight modules for diverse modality inputs, as shown in Table \ref{tab:model_param}.
\subsection{Multi-Task and Universal Training} \label{sec:dataset}
\ours undergoes joint training across various tasks and datasets. Our goal is to assess the capability of a unified model to handle multiple tasks simultaneously. Thus, we refrain from task-specific fine-tuning, despite prior studies demonstrating its potential to enhance task performance.\\
\myparagraph{Various Tasks and Datasets.}
To build a singular unified model for diverse perception and V\&L tasks, we construct an analyzable multi-task benchmark comprising the most representative datasets across five fundamental tasks we previously identified, spanning from image- to pixel-level visual understanding. To enhance the model's adaptability, we augment the benchmark by integrating 27 datasets from 16 publicly accessible data sources, as listed in Table \ref{tab:all_datasets}. \\
\myparagraph{Joint Multi-Task Training.} We jointly train \ours on the above multi-task benchmark by mixing samples from these datasets. As detailed in Table \ref{tab:all_datasets}, to prevent overshadowing tasks with smaller data during joint training and avoid potential performance drops, we uniformly sample from all tasks (1/5), regardless of their data sizes. In universal settings where tasks span multiple domains, sampling inside each task is balanced across scenarios like daily life, indoor, and outdoor. Within these domains, datasets are sampled in proportion to their size. \\
\indent Regarding the learning objective, different tasks require distinct vocabularies. For example, visual grounding uses numerical coordinates, whereas segmentation involves semantic concepts. To tackle this problem, as illustrated in Figure \ref{fig_overview}, we approach all tasks as the next token generation problem using standard CrossEntropy loss, while employing a task-specific vocabulary. This allows for dynamically controlling vocabulary sets, adapting to the unique requirements of each task during both training and inference phases.\\
\myparagraph{Scaling Models.} We adopt a variant of ViT~\cite{dosovitskiy2021an} similar to SAM~\cite{kirillov2023segment}, augmented with six extra transformer layers and text embeddings used in BERT~\cite{devlin2018bert} to improve non-visual modality processing~(refer to Table \ref{tab:ablate_new}). To study the dependence of performance on model scale, we introduce three different sizes of model built up on ViT-B, -L, and -H, with parameters ranging from 131M to 756M, detailed in Table \ref{tab:model_param}. The initial layers inherit parameters pretrained by SAM, while the new layers start with random initialization.
\begin{table*}[t]
    \caption{Results on standard vision-centric benchmarks. ``single-task" refers to models trained on each task separately, while ``multi-task" indicates models trained jointly across all selected benchmarks. ``$\star$'' denotes the model is capable of the task, though no number is reported.  ``-'' means incapability in that specific task. ``$\dag$'' indicates that the generalist model embedded previous task-specific models to enhance performance. \ours stands out as the first generalist model to support all listed vision tasks, delivering competitive outcomes without task-specific adaption. Following ~\cite{li2023uni,chen2023shikra}, some generalist models that only report results with task-specific fine-tuning are not included, \textit{e.g.}, OFA~\cite{wang2022ofa} and X-Decoder\cite{zou2023xdecoder}. We highlight the top-1 entries of one-stage multi-task generalist models and joint training improvements with \textbf{bold} font. Specific module counts exclude non-computational ones, like index-based text tokenizers.}
  \label{tab:all_results}
  \vspace{-4pt}
  \centering
  \resizebox{\textwidth}{!}{
  \begin{tabular}{l|>{\columncolor{mygray}}c>{\columncolor{mygray}}c|c|ccc|ccc|c|cc|c}
    \toprule
    \toprule
    & \multicolumn{2}{>{\columncolor{mygray}}c|}{Specific Modules}&   &  \multicolumn{3}{c|}{Object Detection} & \multicolumn{3}{c|}{Instance Seg} & \multicolumn{1}{c|}{Semantic Seg} & \multicolumn{2}{c|}{Captioning} & \multicolumn{1}{c}{Grounding} \\
    \multirow{-2}{*}{Methods} &Examples & Num & \multirow{-2}{*}{\#Params}  & AP & AP$_{50}$ & AP$_{75}$  & AP & AP$_{50}$ & AP$_{75}$  & mIoU(SS) & BLEU-4 & CIDEr & Acc@0.5 \\
    \midrule
    \multicolumn{14}{l}{\textbf{\textit{Specialist Models}}} \\
    Faster R-CNN-FPN~\cite{ren2015faster} & ResNet,RPN & 5 & 42M    &40.3 & 61.0 &44.0 & - & - & - & - & - & - & - \\
    DETR-DC5~\cite{carion2020detr}        & ResNet,Encoder & 5 & 41M     &43.3 & 63.1 &45.9 & - & - & - & - & - & - & - \\
    Deformable-DETR~\cite{zhu2020deformabledetr}  & ResNet,Encoder & 5 & 40M        &45.4 & 64.7 &49.0 & - & - & - & - & - & - & - \\
    Pix2Seq~\cite{chen2021pix2seq}        & ResNet,Encoder & 3 & 37M       &43.0 & 61.0 & 45.6& - & - & - & - & - & - & -  \\
    Mask R-CNN~\cite{he2017mask}       & ResNet,RPN & 6 & 46M        &41.0 & 61.7 & 44.9 & 37.1 & 58.4 & 40.1 & - & - & - & -  \\
    Polar Mask~\cite{xie2020polarmask} & ResNet,FPN & 5 & 55M       &- & - &- & 30.5 & 52.0 & 31.1 & - & - & - & - \\
    Mask2Former~\cite{cheng2021mask2former}  & ResNet,Decoder & 5 & 44M      &- & - &- & 43.7 & - & - & 47.2 & - & - & -  \\
    VL-T5~\cite{cho2021unifying}        & Faster R-CNN & 3 & 440M   &- & - &- & - & - & - & - & 34.5 & 116.5 & - \\
    UNITER~\cite{chen2020uniter}         & Faster~R-CNN & 4 & 303M         &- & - &- & - & - & - & - & - & - & 81.4 \\
    MDETR~\cite{kamath2021mdetr}       &RoBERTa,DETR& 6& 188M         &- & - &- & - & - & - & - & - & - & 86.8 \\
    \midrule
    \multicolumn{14}{l}{\textbf{\textit{Generalist Models~(Pre-training + MultiTask-Tuning)}}} \\
    UniTab ~\cite{yang2022unitab}          & Encoders & 4 & 185M & - & - &- & - &- & - & - & $\star$ & 115.8 & 88.6 \\
    Pix2Seq v2~\cite{chen2022unified}      & ViT,Decoder & 2 & 132M  & 46.5 &$\star$ &$\star$ & 38.2 &$\star$ & $\star$ & - & 34.9 & $\star$ & $\star$ \\
    Unified-IO$_{XL}$~\cite{lu2023unifiedio} & VQ-VAE & 4 & 2.9B  & - &- &- & - &- & - & $\star$ & $\star$ & 122.3 & $\star$  \\
    Shikra-13B~\cite{chen2023shikra} & ViT,Vicuna & 3 & 13B  & - &- &- & - &- & - & - & $\star$ & 117.5 & 87.8 \\
    \midrule
    \midrule
    \multicolumn{14}{l}{\textbf{\textit{Generalist Models~(MultiTask-Training)}}} \\
    Uni-Perceiver ~\cite{zhu2022uni}       & None & 1 & 124M      & - & - &- & - &- & - & - & 32.0 & $\star$ & $\star$ \\
    Uni-Perceiver-MoE  ~\cite{zhu2022uni-moe}   & None & 1 & 167M  & - & - &- & - &- & - & - & 33.2 & $\star$ & $\star$ \\
    Uni-Perceiver-V2 ~\cite{li2023uni}     & Mask~DINO,Swin & 8 & 308M &\color{gray}58.6$^\dag$ & \color{gray}$\star$ &\color{gray}$\star$ & \color{gray}50.6$^\dag$ &\color{gray}$\star$ & \color{gray}$\star$ & - & 35.4 & 116.9 & $\star$ \\
    VisionLLM-R50 ~\cite{wang2023visionllm}   &Deform-DETR& 6 & 7B  & 44.6 & 64.0 & 48.1 & 25.1 & 50.0 & 22.4 & - & 31.0 & 112.5 & 80.6 \\
    \midrule
    \ours-B$_{\text{single-task}}$         & None &1 & 131M  & 45.1 & 62.7 &49.1 & 31.4 &54.8 & 31.2 & 47.7 & 33.7 & 107.9 & 83.3\\
    \ours-B$_{\text{multi-task}}$       & None &1 & 131M  & 46.7 & 64.2 &50.7 & 31.9 &56.4 & 31.4 & 47.8 & 35.4 & 112.6 & 85.8  \\
    \rowcolor{cyan! 20} \textit{Improvement}$_{~\text{(single}\rightarrow\text{multi)}}$      &  & & & \textbf{\textit{+1.6}} & \textbf{\textit{+1.5}} & \textbf{\textit{+1.6}} & \textbf{\textit{+0.5}} & \textbf{\textit{+1.6}} & \textbf{\textit{+0.2}} & \textbf{\textit{+0.1}} & \textbf{\textit{+1.7}} & \textbf{\textit{+4.7}} & \textbf{\textit{+2.5}} \\
    \ours-L$_{\text{multi-task}}$       & None &1 & 387M  & 51.3 & 69.2 &55.9 & 35.1 &61.4 & 34.7 & 50.6 & 35.7 & 116.0 & 88.4  \\
    \ours-H$_{\text{multi-task}}$       & None &1 & 756M  & \textbf{52.9} & \textbf{71.0} &\textbf{57.8} &\textbf{35.8}  &\textbf{62.6} & \textbf{35.6} & \textbf{52.4} & \textbf{36.2} & \textbf{118.2} & \textbf{89.2} \\
    \toprule
  \end{tabular}
  }
  \vspace{-8pt}
\end{table*}

\section{Experiments}
\subsection{Experimental Settings} \label{sec:datasets}
\myparagraph{Multi-Task Datasets.} To facilitate in-depth analysis and fair evaluation, we built an analyzable multi-task benchmark, choosing one of the most representative datasets for each task. To ensure consistency and enable comparison with VisionLLM~\cite{wang2023visionllm}, we retained the same datasets they used for the four vision-centric tasks: COCO2017~\cite{lin2014coco} for object detection and instance segmentation, COCO Caption~\cite{chen2015microsoft} for image captioning, and the RefCOCO series~\cite{yu2016modeling,mao2016generation} for visual grounding. For the semantic segmentation not included in VisionLLM, we employed the widely used ADE20K dataset~\cite{zhou2017scene}.\\
\myparagraph{Extended Datasets.}  To showcase the universality of our unified framework, we enhanced our multi-task benchmark by integrating more standard and publicly available datasets from vision-language and visual perception~(see \S\ref{sec:dataset}). \\
\myparagraph{Training and Evaluation Details.} To illustrate the flexibility and efficacy of our model, we established three training paradigms: single-task, multi-task, and universal setting. In single-task training, the focus is on optimizing performance on individual benchmarks. Multi-task training, on the other hand, targets the development of a general learner across five selected datasets. Drawing from the insights in Uni-Perceiver v2~\cite{li2023uni}, we adopt an unmixed sampling strategy~(\textit{i.e.}, sampling one task per iteration) for faster and more stable training, However, our framework is also compatible with in-batch mixing strategies~\cite{lu2023unifiedio,zhu2022uni} as suggested by recent studies.~Universal training expands our approach to incorporate 27 comprehensive benchmarks introduced in \S\ref{sec:dataset}.  All models leverage AdamW~\cite{kingma2014adam} optimizer with a cosine annealing schedule, setting the initial learning rate to 0.0002 and weight decay to 0.05. The largest models of the universal setting are trained on 96 NVIDIA A100 GPUs for 320k iterations.\\ 
\indent All experiments are evaluated on the selected datasets using standard protocols and test split. Due to the limited space, more details are in Appendix.
\begin{table*}[t]
    \caption{Zero shot results. ``$\star$'' and ``-'' follow Table~\ref{tab:all_results}. $\dag$ are the performance reproduced based on the mmdetection~\cite{mmdetection}. ``universa'' extends the multi-task setting by including a broader array of datasets, as detailed in \S\ref{sec:dataset}.}
    \label{tab:zero-shot}
    \vspace{-6pt}
    \centering
    \resizebox{1.0\textwidth}{!}{
    \begin{tabular}{l|>{\columncolor{mygray}}c>{\columncolor{mygray}}c|c|c|c|cc|c}
    \toprule
    \toprule
   &\multicolumn{2}{>{\columncolor{mygray}}c|}{Specific Modules}&  &\multicolumn{1}{c|}{Object Detection} & \multicolumn{1}{c|}{Instance Seg} & \multicolumn{2}{c|}{Semantic Seg} & \multicolumn{1}{c}{Captioning} \\
    \multirow{-2}{*}{Methods} &Examples & Num & \multirow{-2}{*}{\#Params} & Cityscapes~\cite{cordts2016cityscapes}  & Cityscapes~\cite{cordts2016cityscapes} & Cityscapes~\cite{cordts2016cityscapes} &
    SUN RGB-D~\cite{song2015sun}& nocaps~\cite{agrawal2019nocaps}\\
    \midrule 
    \multicolumn{8}{l}{\textbf{\textit{Supervised}}} \\
    Faster R-CNN-FPN~\cite{ren2015faster}    &ResNet,RPN& 5 & 42M & 40.3 &-&-&-&-\\
    Mask R-CNN~\cite{he2017mask}             &ResNet,RPN& 6 & 46M & 40.9&36.4&-&-&-\\
    DeepLabV3+~\cite{chen2018deeplabv3+}       &ResNet,Decoder& 3 & 63M & - &-  & 80.9& $\star$&-\\
    Mask2Former~\cite{cheng2021mask2former}  &ResNet,Decoder& 5 & 44M & - &- & 80.4 & $\star$ &-\\
    TokenFusion~\cite{wang2022tokenfusion}   &Segformer,YOLOS & 4 & - &- &- & $\star$ &48.1&- \\
    \midrule
    \midrule
    \multicolumn{9}{l}{\textbf{\textit{Zero-Shot Transfer}}} \\
    GLIP-T~\cite{li2022glip}  &Swin,Dy-Head& 5 & 156M & 28.1$^\dag$& -&-&- & -\\
    Grounding DINO-T~\cite{liu2023groundingdino}  &Swin,DINO &6 &174M & 31.5$^\dag$ & -&-&- & -\\
    BLIP-2~(129M) ~\cite{li2023blip}  &ViT-G,Qformer& 4 & 12.1B & - &- &- &-&\textbf{15.8}\\
    ReCo+~\cite{shin2022reco} &DeiT-SIN & 4 &46M & - &-& 24.2&$\star$ &- \\
    XDecoder-T~\cite{zou2023xdecoder}  &FocalNet,Encoder & 4 &165M &-&16.0 & 47.3&34.5&$\star$ \\
    \midrule
    \ours-B$_{\text{multi-task}}$   &None  &1 &131M & 21.8  &14.3  &34.4  &30.9  &9.2  \\
    \midrule
    \ours-B$_{\text{universal}}$    &None  &1 &131M & 29.1 & 17.9 & 56.2  & 37.5  &10.6\\
    \ours-L$_{\text{universal}}$    &None  &1 &387M & 32.3 & \textbf{20.3} & 58.0  & 39.9  &11.6 \\
    \ours-H$_{\text{universal}}$    &None  &1 &756M & \textbf{34.1} & 18.7 & \textbf{61.8}  & \textbf{42.5}  &12.6\\
    \bottomrule
    \end{tabular}
    }
    \vspace{-12pt}
\end{table*}
\vspace{-20pt}
\subsection{In-distribution Benchmarking} \label{sec:indistbenckmarking}
We evaluate our model's in-distribution performance on various vision-centric tasks, comparing it with both task-specific and advanced generalist models. It relies solely on a stacked multi-layer transformer, adapting to various tasks only through instruction and post-processing changes.\\
\myparagraph{Comparison with Specialist Models.} We compare our single-task model with well-established specialist baselines in Table \ref{tab:all_results}. Our model demonstrates the ability to perform various vision-centric tasks individually within the same framework, narrowing the performance gap with specialized models. It achieves comparable results in most tasks~(\textit{e.g.}, detection: 45.1 vs. 45.4 of Deformable-DETR~\cite{zhu2020deformabledetr}, semantic segmentation: 47.7 vs. 47.2 of Mask2Former~\cite{cheng2021mask2former}), but slightly underperforms in instance segmentation. This is typical for polygon-based methods, which often yield lower results than mask manner. Our model improves by +0.9 against PolarMask~\cite{xie2020polarmask}, a leading polygon-based method. \\ 
\indent Notably, to maintain a universal interface, our method only uses the basic label assignments, without employing the latest enhancement techniques, leaving huge room for performance gains. For example, label assignment used in detection closely mirrors Deformable-DETR~\cite{zhu2020deformabledetr}. Adopting more advanced strategies like DINO's contrastive DeNoising~\cite{zhang2022dino} could further improve our results. \\ 
\myparagraph{Comparison with Generalist Models.} 
Some generalist models~\cite{wang2022ofa,lu2023unifiedio,chen2023shikra,chen2022unified} employ a two-stage training process, initially leveraging large-scale, task-relevant datasets like image-text pairs or diverse perception data, and then undergoing single- or multi-task downstream tuning within the same framework to enhance performance. Our \ours fully embraces the more challenging one-stage joint training, popularized in LLMs, that blends all data for unified modeling followed by direct downstream evaluation, without any task-specific adaptation. \\
\indent Table \ref{tab:all_results} shows that our model not only adeptly manages dense prediction but also outperforms the former leading generalist model, VisionLLM~\cite{wang2023visionllm}, across all tasks, with 50$\times$ fewer parameters and a much simpler framework.\\
\indent Table \ref{tab:all_results},\ref{tab:zero-shot},\ref{tab:few-shot} show that scaling our model greatly improves multitask, zero- and few-shot performance, sometimes even matching supervised approaches.\\
\begin{table*}[t]
    \caption{Few shot results of out-distributed domains. We conduct this experiment based on weights pretrained in the universal stage. ``$\star$'' , ``-'' and $\dag$ follow Table~\ref{tab:zero-shot}.}
    \label{tab:few-shot}
    \vspace{-6pt}
    \centering
    \resizebox{1.0\textwidth}{!}{
    \begin{tabular}{l|>{\columncolor{mygray}}c>{\columncolor{mygray}}c|c|c|c|c|c}
    \toprule
    \toprule
    &\multicolumn{2}{>{\columncolor{mygray}}c|}{Specific Modules}&\multicolumn{1}{c|}{Medical Imaging@mDice}&\multicolumn{2}{c|}{Remote Sensing@mIoU}& \multicolumn{2}{c}{Human Centric@mAP} \\
    \multirow{-2}{*}{Methods} & Examples & Num& DRIVE~\cite{staal2004ridge} &LoveDA~\cite{wang2021loveda}& Potsdam~\cite{potsdam}& WIDERFace~\cite{yang2016wider}&DeepFashion~\cite{liu2016deepfashion} \\
    \midrule 
    \multicolumn{6}{l}{\textbf{\textit{Supervised}}} \\
    U-Net~\cite{ronneberger2015unet} &None &1 &81.4&$\star$&$\star$&-&-\\
    AerialFormer~\cite{yamazaki2023aerialformer}&Encoder,Stem&3  &-&54.1&89.1&-&-\\
    RetinaFace~\cite{deng2020retinaface} &ResNet,FPN &5 &-&-&-&52.3&-\\
    Mask R-CNN~\cite{he2017mask} &ResNet,RPN & 6&-& -&-&$\star$&59.9\\
    \midrule
    \midrule 
    \multicolumn{6}{l}{\textbf{\textit{Few-Shot Transfer}}} \\
    Faster RCNN~\cite{ren2015faster} &ResNet,RPN &4  &  - &- & - & 25.4$^\dag$&14.9$^\dag$\\
    DeepLabV3~\cite{chen2017rethinking}  &ResNet,ASPP & 3&  32.1$^\dag$ &20.3$^\dag$& 24.2$^\dag$& -&- \\
    \midrule
    \ours-B$_{\text{multi-task}}$ & None&1 & 34.3 &24.9 & 19.1 & 17.4 & 23.0 \\
    \midrule
    \ours-B$_{\text{universal}}$ &None &1 &51.1&30.8& 30.6&31.2&38.3\\
    \ours-L$_{\text{universal}}$  &None &1&55.4&34.1 & 37.2 & 33.4&49.3 \\
    \ours-H$_{\text{universal}}$ & None&1 & \textbf{57.9} &\textbf{35.1} & \textbf{43.4}&\textbf{34.0}&\textbf{52.2}\\
    \bottomrule
    \end{tabular}
    }
    \vspace{-12pt}
\end{table*}
\myparagraph{Discussion about multi-task capacity.} 
Table \ref{tab:all_results} reveals that \ours-B$_{\text{multi-task}}$ outperforms \ours-B$_{\text{single-task}}$, showing notable improvements in each task after joint training on five standard datasets. As observed in Table \ref{tab:analysis_multitask}, multi-task training typically boosts performance when tasks share the same capabilities but are less effective otherwise. This pattern is clearly observed in the shared localization ability across detection, visual grounding, and instance segmentation. Conversely, specialized skills, like fine-grained dense prediction in semantic segmentation and polygon-based regression in instance segmentation don't see significant gains from multi-tasking.
\subsection{Out-of-distribution Analysis} \label{sec:generalization}
\myparagraph{Zero-Shot Transfer.} After large-scale multi-task training, \ours is readily assessed on a variety of novel data sources. To demonstrate this capability, we conducted zero-shot evaluations on three established datasets across five configurations, addressing four vision tasks beyond visual grounding. These evaluations span a range of contexts, from indoor environments like SUN RGB-D~\cite{song2015sun}, outdoor scenes such as Cityscapes~\cite{cordts2016cityscapes}, and daily life like nocaps~\cite{agrawal2019nocaps}. We report mIoU and SPICE~\cite{anderson2016spice} for semantic segmentation and captioning, mAP for object detection and instance segmentation.\\
\indent As shown in Table \ref{tab:zero-shot}, our universal models achieve the best results in nearly all tasks. With comparable parameters, \ours-B$_{\text{universal}}$ surpasses X-Decoder~\cite{zou2023xdecoder} on Cityscapes~(+8.9) and SUN RGB-D~(+3.0) on semantic segmentation, and shows similar advantages in instance segmentation and object detection. Scaling the model further enhances its zero-shot capabilities, nearing supervised performance. BLIP-2~\cite{li2023blip} outperforms \ours-H on nocaps, likely attributed to its integration with pretrained language models and extensive training data~(129M). Notably, to our knowledge, \ours is the first generalist model to achieve zero-shot performance across various domains and tasks.\\
\myparagraph{Few-Shot Transfer.} \ours demonstrates rapid adaptation to out-of-distribution data sources. We conducted a comprehensive few-shot evaluation on five datasets in medical imaging (\textit{i.e.}, DRIVE~\cite{staal2004ridge}), remote sensing (\textit{i.e.},LoveDA~\cite{wang2021loveda} and ISPRS~\cite{potsdam}), and human-centric scenarios~(\textit{i.e.}, WIDERFace~\cite{yang2016wider} and DeepFashion~\cite{liu2016deepfashion}). Our approach follows the N-way K-shot~\cite{finn2017model} setting~(\textit{i.e.}, K=5) and directly fine-tune the pre-trained model on support sets~\cite{caelles2017one}. \\
\indent In our segmentation analysis, we choose DeeplabV3 as our baseline, which aligns with the dataset~(\textit{i.e.}, ADE20K) used for training our multi-task variant. We observed that both \ours$_{\text{multi-task}}$ and DeeplabV3 perform poorly in the few-shot setting. However, after large-scale universal training, \ours-B$_{\text{universal}}$ demonstrates significantly improved generalization. This trend is mirrored in detection tasks, underscoring that our universal model structure and training approach greatly enhances generalization capabilities.
\begin{table*}[t]
  \caption{Ablation of modality experts and text conditioning on \ours-B$_{\text{multi-task}}$, using multiple FFN for multimodal learning and image-to-text attention in visual grounding.}
  \label{tab:ablate_experts}
  \vspace{-8pt}
  \centering
  \resizebox{0.9\textwidth}{!}{
  \begin{tabular}{c|c|c|c|c|c|c}
    \toprule
    Modality Experts & Text Conditioning &  Detection@AP & Ins Seg@AP & Sem Seg@mIoU(SS) & Caption@CIDEr & Grounding@Acc(0.5)\\
    \midrule
        & & 46.1 & 31.4 & 47.8 & 111.8 & 78.6  \\
    \checkmark &      & 46.2 & 31.6 & 47.7 & 112.2 & 78.7 \\
    & \checkmark     &46.7 & 31.9 & 47.8 & 112.6 & 85.8 \\
    \toprule
  \end{tabular}
  }
  \vspace{-10pt}
\end{table*}
\begin{table*}[t]
  \caption{Ablation study between encoder-decoder and decoder-only architecture.  }
  \label{tab:ablate_encoder}
  \vspace{-10pt}
  \centering
  \resizebox{0.9\textwidth}{!}{
  \begin{tabular}{l|c|c|c|c|c|c|c}
    \toprule
    Methods & Enc Layer& Dec Layer &  Detection@AP & Ins Seg@AP & Sem Seg@mIoU(SS) & Caption@CIDEr & Grounding @Acc(0.5)\\
    \midrule
    \ours-B$_{\text{multi-task}}$      &12 & 6 &46.3&31.6&46.9&110.8 &84.8 \\
    \ours-B$_{\text{multi-task}}$      &0 & 18 & 46.7 & 31.9 & 47.8 & 112.6 & 85.8  \\
    \toprule
  \end{tabular}
  }
  \vspace{-16pt}
\end{table*}
\subsection{Ablation Study}
\myparagraph{Decoder-only Architecture.} Our model follows the GPT's decoder-only design, though its advantages over encoder-decoder frameworks are not well-explored. We transformed \ours-B's initial 12 layers into an encoder for image and text, excluding target tokens. Table \ref{tab:ablate_encoder} shows that the encoder-decoder paradigm underperforms decoder-only models in all five tasks, particularly in semantic segmentation with a -0.9 drop. This might be due to decoder-only models allocating more layers (18 vs 6) for processing target tokens.
\begin{figure}[t]
    \centering
    \begin{minipage}{0.48\linewidth}
        \centering
        \includegraphics[width=1.\linewidth,height=0.53\linewidth]{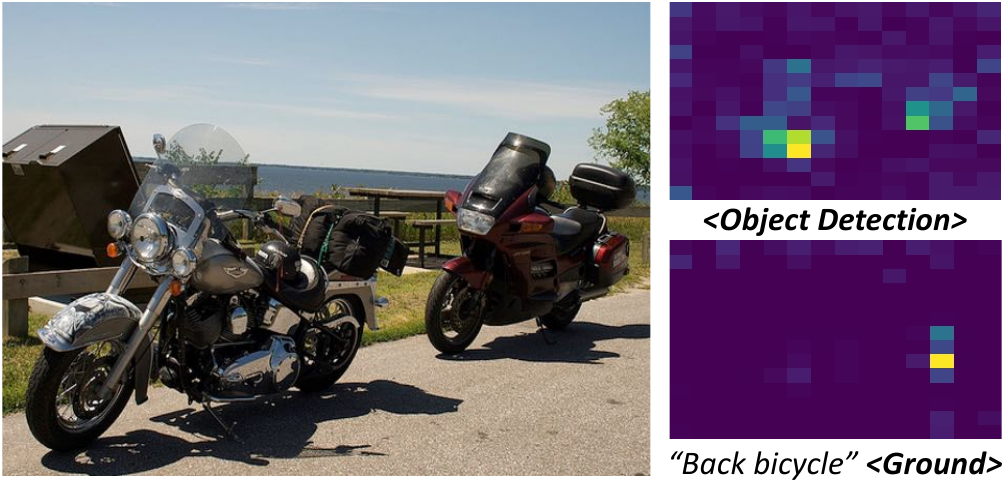}
        \vspace{-20pt}
        \caption{Visualizations on cross-attention between task token and image, with yellower colors indicating higher responses.}
        \label{fig_scalinglaw}
    \end{minipage}
    \hfill
    \begin{minipage}{0.48\linewidth}
        \centering
        \includegraphics[width=1.\linewidth,height=0.53\linewidth]{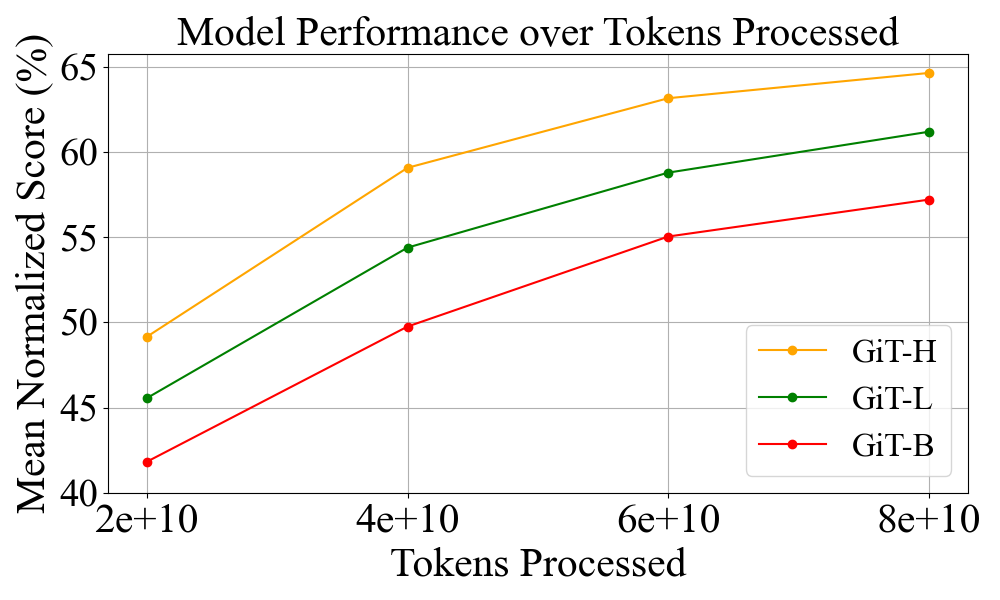}
        \vspace{-20pt}
        \caption{Model size scaling law results. In-distribution performance as a function of tokens processed for 3 model scales. }
        \label{fig_scalinglaw}
    \end{minipage}
    \vspace{-16pt}
\end{figure}

\begin{wraptable}{r}{4.3cm}
        \vspace{-12pt}
        \captionof{table}{Ablation study of new layer on \ours-B$_{\text{single-task}}$.}
        \label{tab:ablate_new}
	\centering
        \resizebox{0.9\linewidth}{!}{
	\begin{tabular}{c|c}
            \toprule
            New Layers & Detection@AP \\
            \midrule
            0 & 40.2 \\
            1 & 42.8 \\
            2 & 43.8 \\
            3 & 44.6 \\
            6 & 45.1 \\
            \toprule
          \end{tabular}
          }
          \vspace{-12pt}
\end{wraptable}

\myparagraph{Number of New Layers.} Table \ref{tab:ablate_new} shows adding just one new layer can significantly boost performance, improving mAP by 2.6, likely due to the difference between image input and language targets. Involving more layers continues to improve results, with gains leveling off after six layers.\\
\myparagraph{Modality Experts.} Although employing multiple FFN as modality experts is a commonly used practice~\cite{bao2022vlmo,zhu2022uni-moe} for multimodal processing, Table \ref{tab:ablate_experts} shows no notable performance gains in our approach, leading us to exclude this design due to its increased parameters and inference latency.\\
\myparagraph{Text Conditioning.} In our visual grounding task with image and text inputs, we enable image-to-text attention during network forwarding. Table \ref{tab:ablate_experts} shows that this method markedly improves performance in a multi-task setting, likely due to its enhanced differentiation between detection and visual grounding tasks. These two tasks function at distinct image scales~(\textit{i.e.}, 1120 and 224), where the former involves identifying multiple boxes, while the latter involves generating a single box guided by text. Moreover, this approach may help the model capture image-text relationships, boosting the ability of instruction-following.\\
\myparagraph{Scaling Law Analysis.} Figure \ref{fig_scalinglaw} presents an in-distribution performance of our universal model against its parameter count, offering insights into the potential enhancements with expanding model capacity. We plot performance progression for three model sizes based on a composite score averaging key metrics from all tasks, showing significant gains with increased scale at a consistent token count.
\section{Conclusion}
In this paper, we introduce \ours, a simple yet powerful vision foundation model that utilizes only a vanilla ViT to integrate diverse visual tasks through a universal language interface. Mirroring multi-task abilities as observed in LLMs, \ours establishes new benchmarks in generalist performance. With training across 27 datasets, \ours becomes the first generalist model to excel in zero- and few-shot tasks across diverse domains using shared parameters, showcasing the foundational role of the multi-layer transformer in computer vision.

\bibliographystyle{splncs04}
\bibliography{main}

\begin{thebibliography}{100}
\providecommand{\url}[1]{\texttt{#1}}
\providecommand{\urlprefix}{URL }
\providecommand{\doi}[1]{https://doi.org/#1}

\bibitem{chowdhery2023palm}
Aakanksha, C., Sharan, N., Jacob, D., Maarten, B., Gaurav, M., Adam, R., Paul, B., Won, C.H., Charles, S., Sebastian, G., et~al.: Palm: Scaling language modeling with pathways. JMLR  (2023)

\bibitem{agrawal2019nocaps}
Agrawal, H., Desai, K., Wang, Y., Chen, X., Jain, R., Johnson, M., Batra, D., Parikh, D., Lee, S., Anderson, P.: Nocaps: Novel object captioning at scale. In: ICCV (2019)

\bibitem{alayrac2022flamingo}
Alayrac, J.B., Donahue, J., Luc, P., Miech, A., Barr, I., Hasson, Y., Lenc, K., Mensch, A., Millican, K., Reynolds, M., et~al.: Flamingo: a visual language model for few-shot learning. In: NeurIPS (2022)

\bibitem{radford2019language}
Alec, R., Jeffrey, W., Rewon, C., David, L., Dario, A., Ilya, S., et~al.: Language models are unsupervised multitask learners. OpenAI blog  (2019)

\bibitem{anderson2016spice}
Anderson, P., Fernando, B., Johnson, M., Gould, S.: Spice: Semantic propositional image caption evaluation. In: ECCV (2016)

\bibitem{bao2022vlmo}
Bao, H., Wang, W., Dong, L., Liu, Q., Mohammed, O.K., Aggarwal, K., Som, S., Piao, S., Wei, F.: Vlmo: Unified vision-language pre-training with mixture-of-modality-experts. In: NeurIPS (2022)

\bibitem{fuyu-8b}
Bavishi, R., Elsen, E., Hawthorne, C., Nye, M., Odena, A., Somani, A., Ta\c{s}\i{}rlar, S.: Introducing our multimodal models (2023), \url{https://www.adept.ai/blog/fuyu-8b}

\bibitem{bommasani2021opportunities}
Bommasani, R., Hudson, D.A., Adeli, E., Altman, R., Arora, S., von Arx, S., Bernstein, M.S., Bohg, J., Bosselut, A., Brunskill, E., et~al.: On the opportunities and risks of foundation models. arXiv preprint arXiv:2108.07258  (2021)

\bibitem{brown2020language}
Brown, T., Mann, B., Ryder, N., Subbiah, M., Kaplan, J.D., Dhariwal, P., Neelakantan, A., Shyam, P., Sastry, G., Askell, A., et~al.: Language models are few-shot learners. In: NeurIPS (2020)

\bibitem{caelles2017one}
Caelles, S., Maninis, K.K., Pont-Tuset, J., Leal-Taix{\'e}, L., Cremers, D., Van~Gool, L.: One-shot video object segmentation. In: CVPR (2017)

\bibitem{caesar2020nuscenes}
Caesar, H., Bankiti, V., Lang, A.H., Vora, S., Liong, V.E., Xu, Q., Krishnan, A., Pan, Y., Baldan, G., Beijbom, O.: nuscenes: A multimodal dataset for autonomous driving. In: CVPR (2020)

\bibitem{caesar2018coco}
Caesar, H., Uijlings, J., Ferrari, V.: Coco-stuff: Thing and stuff classes in context. In: CVPR (2018)

\bibitem{carion2020detr}
Carion, N., Massa, F., Synnaeve, G., Usunier, N., Kirillov, A., Zagoruyko, S.: End-to-end object detection with transformers. In: ECCV (2020)

\bibitem{changpinyo2021conceptual}
Changpinyo, S., Sharma, P., Ding, N., Soricut, R.: Conceptual 12m: Pushing web-scale image-text pre-training to recognize long-tail visual concepts. In: CVPR (2021)

\bibitem{chen2023accelerating}
Chen, C., Borgeaud, S., Irving, G., Lespiau, J.B., Sifre, L., Jumper, J.: Accelerating large language model decoding with speculative sampling. arXiv preprint arXiv:2302.01318  (2023)

\bibitem{mmdetection}
Chen, K., Wang, J., Pang, J., Cao, Y., Xiong, Y., Li, X., Sun, S., Feng, W., Liu, Z., Xu, J., Zhang, Z., Cheng, D., Zhu, C., Cheng, T., Zhao, Q., Li, B., Lu, X., Zhu, R., Wu, Y., Dai, J., Wang, J., Shi, J., Ouyang, W., Loy, C.C., Lin, D.: {MMDetection}: Open mmlab detection toolbox and benchmark. arXiv preprint arXiv:1906.07155  (2019)

\bibitem{chen2023shikra}
Chen, K., Zhang, Z., Zeng, W., Zhang, R., Zhu, F., Zhao, R.: Shikra: Unleashing multimodal llm's referential dialogue magic. arXiv preprint arXiv:2306.15195  (2023)

\bibitem{chen2017rethinking}
Chen, L.C., Papandreou, G., Schroff, F., Adam, H.: Rethinking atrous convolution for semantic image segmentation. CVPR  (2017)

\bibitem{chen2018deeplabv3+}
Chen, L.C., Zhu, Y., Papandreou, G., Schroff, F., Adam, H.: Encoder-decoder with atrous separable convolution for semantic image segmentation. In: ECCV (2018)

\bibitem{chen2020igpt}
Chen, M., Radford, A., Child, R., Wu, J., Jun, H., Luan, D., Sutskever, I.: Generative pretraining from pixels. In: ICML (2020)

\bibitem{chen2021pix2seq}
Chen, T., Saxena, S., Li, L., Fleet, D.J., Hinton, G.: Pix2seq: A language modeling framework for object detection. In: ICLR (2022)

\bibitem{chen2022unified}
Chen, T., Saxena, S., Li, L., Lin, T.Y., Fleet, D.J., Hinton, G.E.: A unified sequence interface for vision tasks. NeurIPS  (2022)

\bibitem{chen2015microsoft}
Chen, X., Fang, H., Lin, T.Y., Vedantam, R., Gupta, S., Doll{\'a}r, P., Zitnick, C.L.: Microsoft coco captions: Data collection and evaluation server. arXiv preprint arXiv:1504.00325  (2015)

\bibitem{chen2020uniter}
Chen, Y.C., Li, L., Yu, L., El~Kholy, A., Ahmed, F., Gan, Z., Cheng, Y., Liu, J.: Uniter: Universal image-text representation learning. In: ECCV (2020)

\bibitem{cheng2021mask2former}
Cheng, B., Misra, I., Schwing, A.G., Kirillov, A., Girdhar, R.: Masked-attention mask transformer for universal image segmentation. In: CVPR (2022)

\bibitem{cho2021unifying}
Cho, J., Lei, J., Tan, H., Bansal, M.: Unifying vision-and-language tasks via text generation. In: ICML (2021)

\bibitem{cordts2016cityscapes}
Cordts, M., Omran, M., Ramos, S., Rehfeld, T., Enzweiler, M., Benenson, R., Franke, U., Roth, S., Schiele, B.: The cityscapes dataset for semantic urban scene understanding. In: CVPR (2016)

\bibitem{instructblip}
Dai, W., Li, J., Li, D., Tiong, A.M.H., Zhao, J., Wang, W., Li, B., Fung, P., Hoi, S.: Instructblip: Towards general-purpose vision-language models with instruction tuning. NeurIPS  (2023)

\bibitem{deng2020retinaface}
Deng, J., Guo, J., Ververas, E., Kotsia, I., Zafeiriou, S.: Retinaface: Single-shot multi-level face localisation in the wild. In: CVPR (2020)

\bibitem{dosovitskiy2021an}
Dosovitskiy, A., Beyer, L., Kolesnikov, A., Weissenborn, D., Zhai, X., Unterthiner, T., Dehghani, M., Minderer, M., Heigold, G., Gelly, S., Uszkoreit, J., Houlsby, N.: An image is worth 16x16 words: Transformers for image recognition at scale. In: ICLR (2021)

\bibitem{everingham2010pascal}
Everingham, M., Van~Gool, L., Williams, C.K., Winn, J., Zisserman, A.: The pascal visual object classes (voc) challenge. IJCV  (2010)

\bibitem{finn2017model}
Finn, C., Abbeel, P., Levine, S.: Model-agnostic meta-learning for fast adaptation of deep networks. In: ICML. PMLR (2017)

\bibitem{gan2020large}
Gan, Z., Chen, Y.C., Li, L., Zhu, C., Cheng, Y., Liu, J.: Large-scale adversarial training for vision-and-language representation learning. NeurIPS  (2020)

\bibitem{girshick2015fastrcnn}
Girshick, R.: Fast r-cnn. In: ICCV (2015)

\bibitem{gupta2019lvis}
Gupta, A., Dollar, P., Girshick, R.: Lvis: A dataset for large vocabulary instance segmentation. In: CVPR (2019)

\bibitem{he2017mask}
He, K., Gkioxari, G., Doll{\'a}r, P., Girshick, R.: Mask r-cnn. In: ICCV (2017)

\bibitem{he2017maskrcnn}
He, K., Gkioxari, G., Doll{\'a}r, P., Girshick, R.: Mask r-cnn. In: ICCV (2017)

\bibitem{he2016deep}
He, K., Zhang, X., Ren, S., Sun, J.: Deep residual learning for image recognition. In: CVPR (2016)

\bibitem{potsdam}
III/4, I.W.: {ISPRS 2D Semantic Labeling Contest}, \url{https://www.isprs.org/education/benchmarks/UrbanSemLab/2d-sem-label-potsdam.aspx}

\bibitem{kamath2021mdetr}
Kamath, A., Singh, M., LeCun, Y., Synnaeve, G., Misra, I., Carion, N.: Mdetr-modulated detection for end-to-end multi-modal understanding. In: ICCV (2021)

\bibitem{karpathy2015deep}
Karpathy, A., Fei-Fei, L.: Deep visual-semantic alignments for generating image descriptions. In: CVPR (2015)

\bibitem{kazemzadeh2014referitgame}
Kazemzadeh, S., Ordonez, V., Matten, M., Berg, T.: Referitgame: Referring to objects in photographs of natural scenes. In: EMNLP (2014)

\bibitem{devlin2018bert}
Kenton, J.D.M.W.C., Toutanova, L.K.: Bert: Pre-training of deep bidirectional transformers for language understanding. In: NAACL-HLT (2019)

\bibitem{kingma2014adam}
Kingma, D.P., Ba, J.: Adam: A method for stochastic optimization  (2015)

\bibitem{kirillov2023segment}
Kirillov, A., Mintun, E., Ravi, N., Mao, H., Rolland, C., Gustafson, L., Xiao, T., Whitehead, S., Berg, A.C., Lo, W.Y., et~al.: Segment anything. In: ICCV (2023)

\bibitem{krishna2017visual}
Krishna, R., Zhu, Y., Groth, O., Johnson, J., Hata, K., Kravitz, J., Chen, S., Kalantidis, Y., Li, L.J., Shamma, D.A., et~al.: Visual genome: Connecting language and vision using crowdsourced dense image annotations. IJCV  (2017)

\bibitem{kuhn1955hungarian}
Kuhn, H.W.: The hungarian method for the assignment problem. NRL  (1955)

\bibitem{kuznetsova2020open}
Kuznetsova, A., Rom, H., Alldrin, N., Uijlings, J., Krasin, I., Pont-Tuset, J., Kamali, S., Popov, S., Malloci, M., Kolesnikov, A., et~al.: The open images dataset v4: Unified image classification, object detection, and visual relationship detection at scale. IJCV  (2020)

\bibitem{li2023uni}
Li, H., Zhu, J., Jiang, X., Zhu, X., Li, H., Yuan, C., Wang, X., Qiao, Y., Wang, X., Wang, W., et~al.: Uni-perceiver v2: A generalist model for large-scale vision and vision-language tasks. In: CVPR (2023)

\bibitem{li2023blip}
Li, J., Li, D., Savarese, S., Hoi, S.: Blip-2: Bootstrapping language-image pre-training with frozen image encoders and large language models. ICML  (2023)

\bibitem{li2022blip}
Li, J., Li, D., Xiong, C., Hoi, S.: Blip: Bootstrapping language-image pre-training for unified vision-language understanding and generation. In: ICML (2022)

\bibitem{li2022glip}
Li, L.H., Zhang, P., Zhang, H., Yang, J., Li, C., Zhong, Y., Wang, L., Yuan, L., Zhang, L., Hwang, J.N., et~al.: Grounded language-image pre-training. In: CVPR (2022)

\bibitem{li2022exploring}
Li, Y., Mao, H., Girshick, R., He, K.: Exploring plain vision transformer backbones for object detection. In: ECCV (2022)

\bibitem{lin2014coco}
Lin, T.Y., Maire, M., Belongie, S., Hays, J., Perona, P., Ramanan, D., Doll{\'a}r, P., Zitnick, C.L.: Microsoft coco: Common objects in context. In: ECCV (2014)

\bibitem{liu2023visual}
Liu, H., Li, C., Wu, Q., Lee, Y.J.: Visual instruction tuning. NeurIPS  (2023)

\bibitem{liu2023groundingdino}
Liu, S., Zeng, Z., Ren, T., Li, F., Zhang, H., Yang, J., Li, C., Yang, J., Su, H., Zhu, J., et~al.: Grounding dino: Marrying dino with grounded pre-training for open-set object detection. arXiv preprint arXiv:2303.05499  (2023)

\bibitem{liu2016deepfashion}
Liu, Z., Luo, P., Qiu, S., Wang, X., Tang, X.: Deepfashion: Powering robust clothes recognition and retrieval with rich annotations. In: CVPR (2016)

\bibitem{long2015fcn}
Long, J., Shelhamer, E., Darrell, T.: Fully convolutional networks for semantic segmentation. In: CVPR (2015)

\bibitem{lu2023unifiedio}
Lu, J., Clark, C., Zellers, R., Mottaghi, R., Kembhavi, A.: {UNIFIED}-{IO}: A unified model for vision, language, and multi-modal tasks. In: ICLR (2023)

\bibitem{mao2016generation}
Mao, J., Huang, J., Toshev, A., Camburu, O., Yuille, A.L., Murphy, K.: Generation and comprehension of unambiguous object descriptions. In: CVPR (2016)

\bibitem{mottaghi2014role}
Mottaghi, R., Chen, X., Liu, X., Cho, N.G., Lee, S.W., Fidler, S., Urtasun, R., Yuille, A.: The role of context for object detection and semantic segmentation in the wild. In: CVPR (2014)

\bibitem{neuhold2017mapillary}
Neuhold, G., Ollmann, T., Rota~Bulo, S., Kontschieder, P.: The mapillary vistas dataset for semantic understanding of street scenes. In: ICCV (2017)

\bibitem{ning2023all}
Ning, J., Li, C., Zhang, Z., Wang, C., Geng, Z., Dai, Q., He, K., Hu, H.: All in tokens: Unifying output space of visual tasks via soft token. In: ICCV (2023)

\bibitem{openAI2023chatgpt}
OpenAI: Chatgpt (2022), \url{https://openai.com/blog/chatgpt}

\bibitem{openAI2023gpt4}
OpenAI: Gpt-4 technical report (2023)

\bibitem{ordonez2011im2text}
Ordonez, V., Kulkarni, G., Berg, T.: Im2text: Describing images using 1 million captioned photographs. NeurIPS  \textbf{24} (2011)

\bibitem{ouyang2022instrucGPT}
Ouyang, L., Wu, J., Jiang, X., Almeida, D., Wainwright, C., Mishkin, P., Zhang, C., Agarwal, S., Slama, K., Ray, A., et~al.: Training language models to follow instructions with human feedback. NeurIPS  \textbf{35} (2022)

\bibitem{plummer2015flickr30k}
Plummer, B.A., Wang, L., Cervantes, C.M., Caicedo, J.C., Hockenmaier, J., Lazebnik, S.: Flickr30k entities: Collecting region-to-phrase correspondences for richer image-to-sentence models. In: ICCV (2015)

\bibitem{radford2018improving}
Radford, A., Narasimhan, K., Salimans, T., Sutskever, I., et~al.: Improving language understanding by generative pre-training  (2018)

\bibitem{raffel2020t5}
Raffel, C., Shazeer, N., Roberts, A., Lee, K., Narang, S., Matena, M., Zhou, Y., Li, W., Liu, P.J.: Exploring the limits of transfer learning with a unified text-to-text transformer. JMLR  (2020)

\bibitem{ramesh2021zero}
Ramesh, A., Pavlov, M., Goh, G., Gray, S., Voss, C., Radford, A., Chen, M., Sutskever, I.: Zero-shot text-to-image generation. In: ICML (2021)

\bibitem{reed2022generalist}
Reed, S., Zolna, K., Parisotto, E., Colmenarejo, S.G., Novikov, A., Barth-Maron, G., Gimenez, M., Sulsky, Y., Kay, J., Springenberg, J.T., et~al.: A generalist agent. TMLR  (2022)

\bibitem{ren2015faster}
Ren, S., He, K., Girshick, R., Sun, J.: Faster r-cnn: Towards real-time object detection with region proposal networks. NeurIPS  (2015)

\bibitem{ronneberger2015unet}
Ronneberger, O., Fischer, P., Brox, T.: U-net: Convolutional networks for biomedical image segmentation. In: MICCAI (2015)

\bibitem{shao2019objects365}
Shao, S., Li, Z., Zhang, T., Peng, C., Yu, G., Zhang, X., Li, J., Sun, J.: Objects365: A large-scale, high-quality dataset for object detection. In: ICCV (2019)

\bibitem{sharma2018conceptual}
Sharma, P., Ding, N., Goodman, S., Soricut, R.: Conceptual captions: A cleaned, hypernymed, image alt-text dataset for automatic image captioning. In: ACL (2018)

\bibitem{shin2022reco}
Shin, G., Xie, W., Albanie, S.: Reco: Retrieve and co-segment for zero-shot transfer. In: NeurIPS (2022)

\bibitem{song2015sun}
Song, S., Lichtenberg, S.P., Xiao, J.: Sun rgb-d: A rgb-d scene understanding benchmark suite. In: CVPR (2015)

\bibitem{staal2004ridge}
Staal, J., Abr{\`a}moff, M.D., Niemeijer, M., Viergever, M.A., Van~Ginneken, B.: Ridge-based vessel segmentation in color images of the retina. TMI  (2004)

\bibitem{alpaca}
Taori, R., Gulrajani, I., Zhang, T., Dubois, Y., Li, X., Guestrin, C., Liang, P., Hashimoto, T.B.: Stanford alpaca: An instruction-following llama model. \url{https://github.com/tatsu-lab/stanford_alpaca} (2023)

\bibitem{team2023gemini}
Team, G., Anil, R., Borgeaud, S., Wu, Y., Alayrac, J.B., Yu, J., Soricut, R., Schalkwyk, J., Dai, A.M., Hauth, A., et~al.: Gemini: a family of highly capable multimodal models. arXiv preprint arXiv:2312.11805  (2023)

\bibitem{touvron2023llama}
Touvron, H., Lavril, T., Izacard, G., Martinet, X., Lachaux, M.A., Lacroix, T., Rozi{\`e}re, B., Goyal, N., Hambro, E., Azhar, F., et~al.: Llama: Open and efficient foundation language models. arXiv preprint arXiv:2302.13971  (2023)

\bibitem{van2017vqvae}
Van Den~Oord, A., Vinyals, O., et~al.: Neural discrete representation learning. In: NeurIPS (2017)

\bibitem{vaswani2017attention}
Vaswani, A., Shazeer, N., Parmar, N., Uszkoreit, J., Jones, L., Gomez, A.N., Kaiser, {\L}., Polosukhin, I.: Attention is all you need. In: NeurIPS (2017)

\bibitem{wang2023dsvt}
Wang, H., Shi, C., Shi, S., Lei, M., Wang, S., He, D., Schiele, B., Wang, L.: Dsvt: Dynamic sparse voxel transformer with rotated sets. In: CVPR (2023)

\bibitem{wang2023unitr}
Wang, H., Tang, H., Shi, S., Li, A., Li, Z., Schiele, B., Wang, L.: Unitr: A unified and efficient multi-modal transformer for bird's-eye-view representation. In: ICCV (2023)

\bibitem{wang2021loveda}
Wang, J., Zheng, Z., Ma, A., Lu, X., Zhong, Y.: Loveda: A remote sensing land-cover dataset for domain adaptive semantic segmentation. In: NeurIPS (2021)

\bibitem{wang2022ofa}
Wang, P., Yang, A., Men, R., Lin, J., Bai, S., Li, Z., Ma, J., Zhou, C., Zhou, J., Yang, H.: Ofa: Unifying architectures, tasks, and modalities through a simple sequence-to-sequence learning framework. In: ICML (2022)

\bibitem{wang2023visionllm}
Wang, W., Chen, Z., Chen, X., Wu, J., Zhu, X., Zeng, G., Luo, P., Lu, T., Zhou, J., Qiao, Y., et~al.: Visionllm: Large language model is also an open-ended decoder for vision-centric tasks. NeurIPS  (2023)

\bibitem{wang2022image}
Wang, W., Bao, H., Dong, L., Bjorck, J., Peng, Z., Liu, Q., Aggarwal, K., Mohammed, O.K., Singhal, S., Som, S., et~al.: Image as a foreign language: Beit pretraining for all vision and vision-language tasks. CVPR  (2023)

\bibitem{wang2022tokenfusion}
Wang, Y., Chen, X., Cao, L., Huang, W., Sun, F., Wang, Y.: Multimodal token fusion for vision transformers. In: CVPR (2022)

\bibitem{wu2016google}
Wu, Y., Schuster, M., Chen, Z., Le, Q.V., Norouzi, M., Macherey, W., Krikun, M., Cao, Y., Gao, Q., Macherey, K., et~al.: Google's neural machine translation system: Bridging the gap between human and machine translation. arXiv preprint arXiv:1609.08144  (2016)

\bibitem{xie2020polarmask}
Xie, E., Sun, P., Song, X., Wang, W., Liu, X., Liang, D., Shen, C., Luo, P.: Polarmask: Single shot instance segmentation with polar representation. In: CVPR (2020)

\bibitem{xu2019explicit}
Xu, W., Wang, H., Qi, F., Lu, C.: Explicit shape encoding for real-time instance segmentation. In: ICCV (2019)

\bibitem{yamazaki2023aerialformer}
Yamazaki, K., Hanyu, T., Tran, M., Garcia, A., Tran, A., McCann, R., Liao, H., Rainwater, C., Adkins, M., Molthan, A., et~al.: Aerialformer: Multi-resolution transformer for aerial image segmentation. arXiv preprint arXiv:2306.06842  (2023)

\bibitem{yang2016wider}
Yang, S., Luo, P., Loy, C.C., Tang, X.: Wider face: A face detection benchmark. In: CVPR (2016)

\bibitem{yang2022unitab}
Yang, Z., Gan, Z., Wang, J., Hu, X., Ahmed, F., Liu, Z., Lu, Y., Wang, L.: Unitab: Unifying text and box outputs for grounded vision-language modeling. In: ECCV (2022)

\bibitem{you2023ferret}
You, H., Zhang, H., Gan, Z., Du, X., Zhang, B., Wang, Z., Cao, L., Chang, S.F., Yang, Y.: Ferret: Refer and ground anything anywhere at any granularity. In: ICLR (2024)

\bibitem{yu2020bdd100k}
Yu, F., Chen, H., Wang, X., Xian, W., Chen, Y., Liu, F., Madhavan, V., Darrell, T.: Bdd100k: A diverse driving dataset for heterogeneous multitask learning. In: CVPR (2020)

\bibitem{yu2016modeling}
Yu, L., Poirson, P., Yang, S., Berg, A.C., Berg, T.L.: Modeling context in referring expressions. In: ECCV. Springer (2016)

\bibitem{zhang2022dino}
Zhang, H., Li, F., Liu, S., Zhang, L., Su, H., Zhu, J., Ni, L.M., Shum, H.Y.: Dino: Detr with improved denoising anchor boxes for end-to-end object detection. In: ICLR (2022)

\bibitem{zhang2022opt}
Zhang, S., Roller, S., Goyal, N., Artetxe, M., Chen, M., Chen, S., Dewan, C., Diab, M., Li, X., Lin, X.V., et~al.: Opt: Open pre-trained transformer language models. arXiv preprint arXiv:2205.01068  (2022)

\bibitem{zhou2017scene}
Zhou, B., Zhao, H., Puig, X., Fidler, S., Barriuso, A., Torralba, A.: Scene parsing through ade20k dataset. In: CVPR (2017)

\bibitem{zhu2023minigpt}
Zhu, D., Chen, J., Shen, X., Li, X., Elhoseiny, M.: Minigpt-4: Enhancing vision-language understanding with advanced large language models. arXiv preprint arXiv:2304.10592  (2023)

\bibitem{zhu2022uni-moe}
Zhu, J., Zhu, X., Wang, W., Wang, X., Li, H., Wang, X., Dai, J.: Uni-perceiver-moe: Learning sparse generalist models with conditional moes. NeurIPS  (2022)

\bibitem{zhu2020deformabledetr}
Zhu, X., Su, W., Lu, L., Li, B., Wang, X., Dai, J.: Deformable detr: Deformable transformers for end-to-end object detection. ICLR  (2020)

\bibitem{zhu2022uni}
Zhu, X., Zhu, J., Li, H., Wu, X., Li, H., Wang, X., Dai, J.: Uni-perceiver: Pre-training unified architecture for generic perception for zero-shot and few-shot tasks. In: CVPR (2022)

\bibitem{zou2023xdecoder}
Zou, X., Dou, Z.Y., Yang, J., Gan, Z., Li, L., Li, C., Dai, X., Behl, H., Wang, J., Yuan, L., et~al.: Generalized decoding for pixel, image, and language. In: CVPR (2023)

\end{thebibliography}

\newpage
\appendix
In our supplementary, we provide detailed information including model design specifics in \S \ref{sec:arc}, dataset summaries in \S \ref{sec:datasets}, along with in-depth training, inference and evaluation procedures in \S \ref{sec:train} and \ref{sec:inference}. Additional ablation experiments are included in \S \ref{sec:ablate}. \S \ref{sec:specifc} details specific modules used in comparative methods. Qualitative results across different datasets and tasks are in \S\ref{sec:vis}. Lastly, limitations, negative societal impacts, and a comparison with Fuyu-8B are in \S\ref{sec:discussion}.

\section{Implementation details} \label{sec:arc}
\begin{figure}[t]
    \centering
    \begin{minipage}{0.44\linewidth}
        \centering
        \vspace{-10pt}
        \begin{center}
        \includegraphics[width=0.9\linewidth]{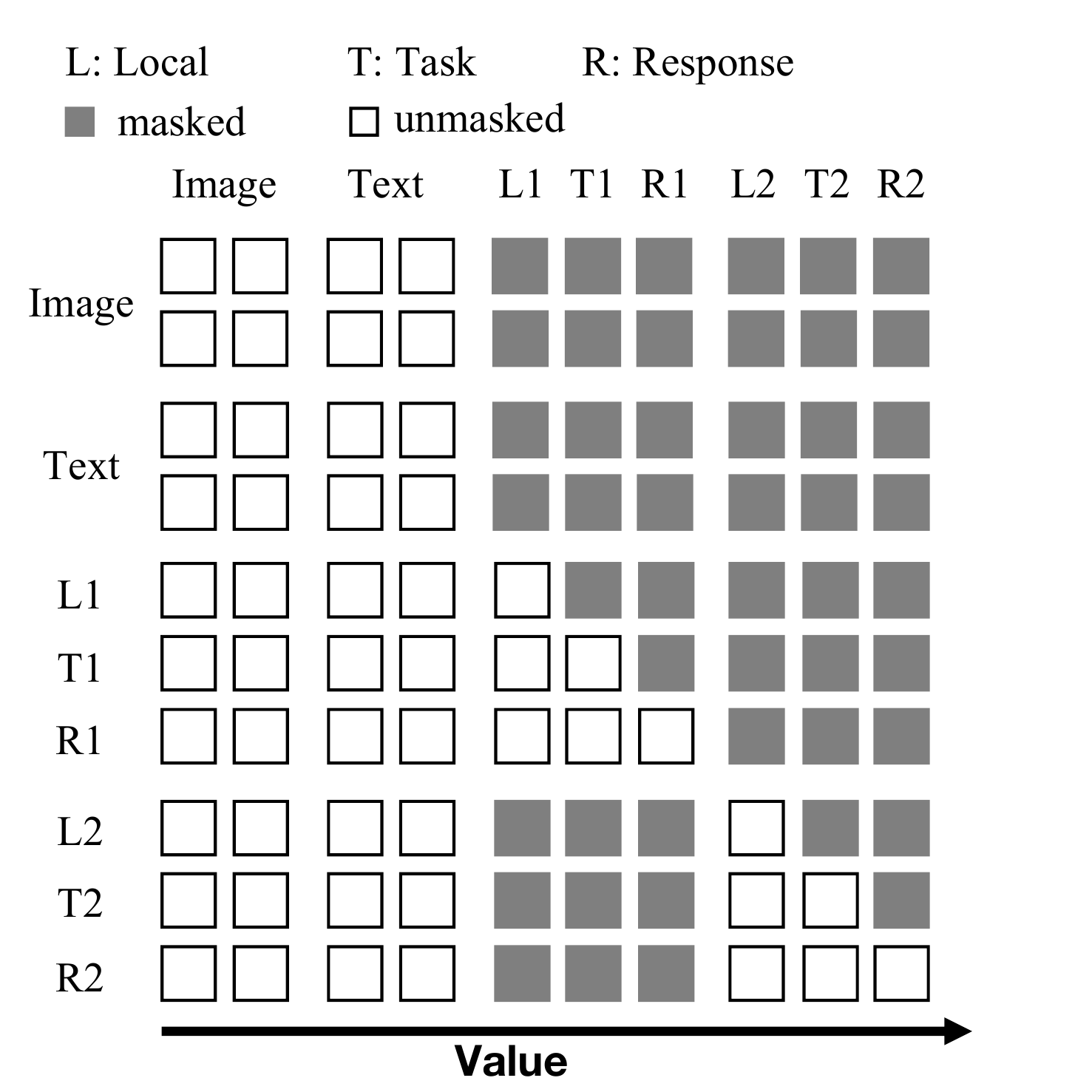}
        \end{center}
        \vspace{-20pt}
        \caption{Attention mask visualization.}
        \vspace{-12pt}
        \label{fig_attn_mask}
    \end{minipage}
    \hfill
    \begin{minipage}{0.55\linewidth}
        \centering
        \vspace{14pt}
        \begin{center}
        \includegraphics[width=0.9\linewidth]{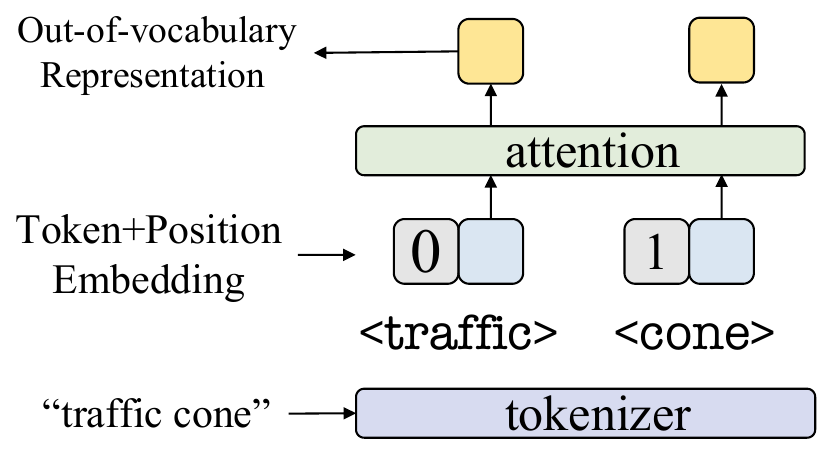}
        \end{center}
        \vspace{-4pt}
        \caption{Out-of-vocabulary representation.}
        \label{fig_out-of-vocabulary}
        \vspace{-16pt}
    \end{minipage}
    \vspace{-4pt}
\end{figure}

\myparagraph{Window Attention.}
Our window attention is adapted from the SAM~\cite{kirillov2023segment} variant of ViT~\cite{dosovitskiy2021an}. Following SAM, after patch embedding, images are downsampled by a factor of 16, and windows are defined with a size of 14$\times$14. The primary distinction from the original lies in how we handle multi-track local observations and responses in the parallel training stage, such as grid-wise prompts (\textit{i.e.}, local image token, task identifier) and their outputs.
To manage these multi-track elements, we merge them into a sequence and append them after the shared observation. Consequently, the input to window attention consists of multiple parts, requiring a customized attention mask to ensure grid independence while enabling autoregressive prediction, as detailed in Figure \ref{fig_attn_mask}. Within each subprocess group (\textit{i.e.}, those associated with the same grid), interactions are left-to-right unidirectional attention. Moreover, tokens belonging to different subprocesses are isolated, preventing them from accessing each other's information.\\
\myparagraph{Global Attention.}
In tasks that require object- and pixel-level analysis, the large number of local predictions creates significant memory and computational burdens, especially in global attention layers, where processing attention across all grid points can be unnecessary and inefficient. Therefore, for such tasks, we have optimized the global attention layer to focus only on the shared global observations~(\textit{i.e.}, input image and text), eliminating the need to compute targets for each grid. Table \ref{tab:global_only} shows that this strategy slightly impacts performance but greatly decreases computation time. However, in captioning and visual grounding with a 224 image size, which involves only one window and a single global response, this optimization is unnecessary.
\begin{table}
  \centering
  \caption{Performance of semantic segmentation by single-task training with our accelerated global attention. It significantly reduces the computational cost with slight performance drops.}
  \label{tab:global_only}
  \vspace{-6pt}
  \resizebox{0.4\linewidth}{!}{
  \begin{tabular}{c|c|c}
    \toprule
    Global Attention & mIoU & Training Time \\
    \midrule
     Normal & 47.9& 51h \\
    Accelerated  & 47.7 & 35h \\
    \bottomrule
  \end{tabular}
  }
  \vspace{-2pt}
\end{table}

\myparagraph{Out-of-vocabulary Representation.}
We encode multi-piece out-of-vocabulary concepts to a single token. This is achieved through a streamlined approach that utilizes only one attention layer combined with absolute positional encoding. As shown in Figure \ref{fig_out-of-vocabulary}, ``traffic cone'' is tokenized as \texttt{$<$\text{traffic}$>$}\texttt{$<$\text{cone}$>$}. The corresponding text embeddings, augmented with positional encoding, are input into the attention layer, allowing each word to interact with the rest. We select the first output token as the final representation for multi-word concepts like ``traffic cone''. For single-word concepts, we use the original text embedding directly.\\
\myparagraph{Background Representation.}
Given that each dataset contains distinct positive and negative classes, utilizing text labels like \texttt{$<$\text{background}$>$} to denote negative classes could lead to ambiguity when training across multiple datasets. Therefore, we employed a unique encoding approach for the background class,
\begin{equation}
    \begin{aligned}
    \mathcal{F}_{\text{background}} = - \sum_{i=0}^{N-1} \mathcal{F}_i / N
    \end{aligned}
\vspace{-4pt}
\end{equation}
where $\mathcal{F}_{i}$ is the representation of $i$-th positive class and $N$ denotes the total number of categories. 
This approach makes the cosine similarity between tokens of a positive class and those assigned to the background class typically negative. Its superior performance in zero-shot scenarios highlights its effectiveness.\\
\myparagraph{Resolution and Coordinate Discretization.}
For our experiments, we use different image resolutions tailored to specific tasks: 1120 $\times$ 1120 pixels for object detection and instance segmentation, 672 $\times$ 672 pixels for semantic segmentation, and 224 $\times$ 224 pixels for image captioning and visual grounding. To encode spatial positions as discrete tokens, we discretize the image coordinates into a set number of intervals. Specifically, we determine the number of these intervals to be double the resolution of the input image. For instance, with an input image of 224 $\times$ 224 pixels, we divide the coordinate space into 448 discrete intervals.
\section{Extended Datasets} \label{sec:datasets}
\begin{table}
  \centering
  \caption{
Universal training dataset details. Columns from left to right indicate dataset size, proportion to total data, assigned group number, and sampling weight. Weights are evenly distributed across the tasks. Different scenarios within each task (\textit{e.g.}, daily life, autonomous driving) create individual groups with equal weights. Sampling weights in groups are set based on dataset sizes.}
  \label{tab:all_datasets}
  \resizebox{0.8\linewidth}{!}{
  \begin{tabular}{l|c|c|c|c}
    \toprule
    Dataset  & Size & Percent~(\%) & Group ID & Weight~(\%) \\
    \midrule
    \textbf{Object Detection} & 3.8M&22.55 &- &20.00\\
    Objects365~\cite{shao2019objects365} & 1.7M & 9.98 & 0 &3.22 \\
    OpenImages~\cite{kuznetsova2020open}  & 1.7M & 9.98 &0  & 3.22 \\
    LVIS~\cite{gupta2019lvis}  & 164K & 0.96 & 0 & 0.23 \\
    nuImages~\cite{caesar2020nuscenes} &93K & 0.55 & 1& 6.66\\
    Pascal VOC 2007~\cite{everingham2010pascal} &10K& 0.06 & 2& 0.37\\
    Pascal VOC 2012~\cite{everingham2010pascal} & 11K &  0.06&2  &0.22  \\
    COCO 2017~\cite{lin2014coco}  & 164K & 0.96  & 2 & 6.07 \\
    \textbf{Instance Segmentation}&1.4M& 8.34&- &20.00\\
    LVIS~\cite{gupta2019lvis}  & 164K &0.96 &3 &0.76 \\
    OpenImages~\cite{kuznetsova2020open}  & 1M &5.87 &3  &5.90 \\
    nuImages~\cite{caesar2020nuscenes}  & 93K & 0.55&4 &6.66 \\
    COCO 2017~\cite{lin2014coco} & 164K & 0.96&5 &6.66\\

    \textbf{Semantic Segmentation}&322K&1.89& -& 20.00\\
    COCO-Stuff~\cite{caesar2018coco} & 164K &0.96 &6&6.28\\

    Pascal Context~\cite{mottaghi2014role} & 10K &0.06&6&0.38\\
    nuImages~\cite{caesar2020nuscenes} & 93K &0.55&7&4.84\\

    BDD100K~\cite{yu2020bdd100k} & 10K &0.06&7&0.52 \\
    Mapillary Vistas~\cite{neuhold2017mapillary} &25K&0.15 &7 &1.30 \\
    ADE20K~\cite{zhou2017scene} & 20K & 0.12&8 &6.67 \\

    \textbf{Image Caption}&11.3M&66.54&-&20.00\\
    CC3M~\cite{sharma2018conceptual} & 1.8M &10.57&9&1.74\\
    CC12M ~\cite{changpinyo2021conceptual}& 7.8M &45.79&9&6.96\\
    SBU Captions~\cite{ordonez2011im2text} & 800K &4.70&9&0.58\\
    Visual Genome ~\cite{krishna2017visual}& 770K &4.52& 9&0.71 \\
    COCO Caption~\cite{chen2015microsoft} & 164K &0.96&10&10.00\\
    \textbf{Visual Grounding }&115K&0.68&-&20.00\\
    RefCOCO~\cite{kazemzadeh2014referitgame} & 20K &0.12&11& 4.00 \\
    RefCOCO+~\cite{kazemzadeh2014referitgame} & 20K &0.12&11& 4.00\\
    RefCOCOg~\cite{mao2016generation} & 25K &0.15&11& 4.00 \\
    RefCLEF~\cite{kazemzadeh2014referitgame} & 20K &0.12&12&4.00\\
    Flickr30K~\cite{plummer2015flickr30k} & 30K &0.18&13&4.00\\
      \textbf{All} & 17M &100 & - & 100 \\
    \bottomrule
  
  \end{tabular}
  }
  \vspace{-8pt}
\end{table}
\subsection{In-distribution Datasets}
During universal training, a total of 27 datasets from 16
publicly accessible data sources are used, with sizes and weights detailed in Table \ref{tab:all_datasets}. Note that the actual quantities in web-sourced caption datasets (CC3M~\cite{sharma2018conceptual}, CC12M~\cite{changpinyo2021conceptual}, SBU Captions~\cite{ordonez2011im2text}) are fewer than the original number reported due to inactive links.\\
\myparagraph{COCO.} The MS COCO dataset, or Microsoft Common Objects in Context~\cite{lin2014coco}, is a comprehensive dataset for object detection, segmentation, key-point detection, and captioning. It includes over 330K images, with annotations for more than 220K, featuring 1.5 million objects across 80 categories. Each image has five sentence descriptions and 250K pedestrians are annotated with keypoints. The initial release in 2014 has 164K images in training (83K), validation (41K), and test (41K) sets. In 2017, the training/validation split changed to 118K/5K. \\
\myparagraph{Objects365.} Objects365~\cite{shao2019objects365} is a vast object detection dataset, comprising 365 object categories and boasting over 2 million training images along with 30 million annotated bounding boxes. This dataset presents diverse objects in different scenarios, providing a robust benchmark for challenging object detection tasks. \\
\myparagraph{OpenImages.} 
Open Images~\cite{kuznetsova2020open} is a dataset with about 9 million images, each annotated with image-level labels, object bounding boxes, segmentation masks, visual relationships, localized narratives, and point-level labels. Covering 20,638 image-level labels, 600 object classes with 16 million bounding boxes, and 2.8 million segmentation masks, it stands as a valuable resource in computer vision. \\
\myparagraph{LVIS.} LVIS~\cite{gupta2019lvis} (Large Vocabulary Instance Segmentation) is a dataset tailored for instance segmentation tasks, providing approximately 2 million high-quality segmentation masks across over 1000 entry-level object categories within a dataset of 164,000 images. This dataset was created to tackle the Zipf distribution commonly observed in natural images, making it an invaluable resource for researchers and developers working on instance segmentation tasks dealing with a large vocabulary of objects.\\
\myparagraph{Pascal VOC 2007.} The Pascal VOC 2007~\cite{everingham2010pascal} dataset serves as a crucial resource for real-world object recognition, featuring 20 object classes. With 9,963 photos and 24,640 labeled samples, thoughtfully split for balanced training/validation and testing, it stands as a versatile dataset supporting various tasks, including classification, detection, segmentation, and person layout. \\
\myparagraph{Pascal VOC 2012.} Pascal VOC 2012~\cite{everingham2010pascal} is a valuable dataset for recognizing objects in real-world settings. It encompasses 20 object classes and includes 11,530 images with 27,450 ROI-tagged objects and 6,929 segmentations, serving as a prominent benchmark in computer vision. \\
\myparagraph{nuImages.}
The nuImages~\cite{caesar2020nuscenes} dataset complements the nuScenes~\cite{caesar2020nuscenes} for autonomous driving by providing 93,000 2D annotated images, with 1.2 million camera images from past and future timestamps. It is part of the nuScenes ecosystem and focuses on panoptic and multi-annotation aspects. The dataset covers various driving scenarios, including diverse conditions such as rain, snow, and night. It also offers temporal dynamics with 2 Hz spaced images. The annotations encompass 800,000 foreground objects with instance masks and 100,000 semantic segmentation masks. \\
\myparagraph{ADE20K.}
The ADE20K~\cite{zhou2017scene} semantic segmentation dataset comprises 20,000 scene-centric images meticulously annotated at the pixel level for both objects and object parts. Encompassing 150 semantic categories, it includes items like sky, road, and specific objects such as person, car, and bed. The dataset is divided into 20,210 training, 2,000 validation, and 3,000 testing images. \\
\myparagraph{COCO-Stuff.}
The COCO-stuff~\cite{caesar2018coco} dataset holds significance for diverse scene understanding tasks, such as semantic segmentation, object detection, and image captioning. Derived by augmenting the original COCO dataset, which initially prioritized object annotations, it addresses the oversight of stuff annotations. Spanning 164,000 images, the COCO-stuff dataset includes 172 categories, incorporating 80 things, 91 stuff, and 1 unlabeled class. \\
\myparagraph{Pascal Context.}
The PASCAL Context~\cite{mottaghi2014role} dataset extends the PASCAL VOC 2010~\cite{everingham2010pascal} detection challenge by providing pixel-wise labels for all training images. Encompassing over 400 classes, which include the original 20 classes from PASCAL VOC segmentation, these classes are categorized into objects, stuff, and hybrids. To address the sparsity of many object categories, a common practice involves using a subset of 59 frequently occurring classes.\\
\myparagraph{BDD100K.} BDD100K~\cite{yu2020bdd100k} is a large dataset with 100K videos, providing over 1,000 hours of driving experience and 100 million frames. It includes annotations for road objects, lane markings, drivable areas, and detailed instance segmentation. For road object detection and drivable area segmentation challenges, there are 70,000 training and 10,000 validation images. For full-frame semantic segmentation, there are 7,000 training and 1,000 validation images. \\ 
\myparagraph{Mapillary Vistas.}
Mapillary Vistas~\cite{neuhold2017mapillary} is a large-scale street-level image dataset with 25,000 high-resolution images. Featuring annotations for 66 object categories, including instance-specific labels for 37 classes, it adopts a dense and fine-grained annotation style using polygons. The dataset primarily focuses on semantic image segmentation and instance-specific image segmentation, aiming to advance visual road-scene understanding. \\
\myparagraph{CC3M.}
Conceptual Captions, known as CC3M~\cite{sharma2018conceptual}, features an extensive collection of around 3.3 million images, each meticulously paired with descriptive captions. Extracted from Alt-text HTML attributes associated with web images, these captions undergo an automated pipeline for quality assurance. This makes the dataset highly versatile, catering to a diverse range of natural language processing and image understanding tasks. \\
\myparagraph{CC12M.}
Conceptual 12M~\cite{changpinyo2021conceptual} (CC12M) is a dataset specifically created for vision-and-language pre-training. It consists of a substantial 12 million image-text pairs. Unlike some other datasets with restrictive requirements, CC12M relaxes its data collection pipeline to enhance dataset scale and diversity. It has been shown to provide state-of-the-art results in vision-and-language tasks, particularly in long-tail visual recognition, making it a valuable resource for research and development in this field. \\
\myparagraph{SBU Captions.}
The SBU Captions dataset~\cite{ordonez2011im2text} is a collection of 1 million images and their associated captions sourced from Flickr, primarily used for training image captioning models. It provides diverse real-world images and textual descriptions, serving as a valuable resource for research in computer vision and natural language processing. \\
\myparagraph{Visual Genome.}
Visual Genome~\cite{krishna2017visual} is a comprehensive dataset with 108,077 images, richly annotated with 5.4 million region descriptions, 1.7 million visual question answers, 3.8 million object instances, 2.8 million attributes, and 2.3 million relationships. This dataset is designed to provide detailed information about images, including objects, attributes, and the relationships between them. \\
\myparagraph{COCO Caption.}
COCO Captions~\cite{chen2015microsoft} consists of 1.5 million captions for 330,000 images, with five captions for each image in the training and validation sets. The ``Karpathy split'', a widely used subset of this dataset created by Andrej Karpathy, involves merging the train and val sets from the raw dataset, creating a new validation set by selecting 5,000 images from the original val set, and an additional 5,000 images are used to form a test set. \\
\myparagraph{RefCOCO.}
The RefCOCO~\cite{kazemzadeh2014referitgame}, RefCOCO+~\cite{kazemzadeh2014referitgame}, and RefCOCOg~\cite{mao2016generation} datasets were generated through the ReferitGame, a two-player game where one participant describes a segmented object in an image using natural language, and the other participant identifies the correct object. In RefCOCO, there are no language restrictions on referring expressions, whereas in RefCOCO+, location words are prohibited. These datasets concentrate on appearance-based descriptions, such as ``the man in the yellow polka-dotted shirt," rather than perspective-dependent ones. RefCOCO comprises 142,209 referring expressions for 50,000 objects in 19,994 images, and RefCOCO+ contains 141,564 expressions for 49,856 objects in 19,992 images.\\
\myparagraph{RefCLEF.}
RefCLEF~\cite{kazemzadeh2014referitgame}, also known as ReferIt, consists of 20,000 images sourced from the IAPR TC-12 dataset, accompanied by segmented image regions from the SAIAPR-12 dataset. The dataset is evenly split into two sections: one with 10,000 images designated for training and validation, and another with 10,000 images for testing. The training and validation portion includes a total of 59,976 entries, each consisting of an image, a bounding box, and a description. Test set is slightly larger, featuring 60,105 entries with the same type of data.\\
\myparagraph{Flickr30K.}
Flickr30K~\cite{plummer2015flickr30k} is a widely recognized dataset used for sentence-based image descriptions. It features 31,783 images depicting everyday activities and events, each accompanied by a descriptive caption. This dataset serves as a standard benchmark for studying the relationship between linguistic expressions and visual media.
\subsection{Out-distribution Datasets}
\myparagraph{Cityscapes.}
Cityscapes~\cite{cordts2016cityscapes} is a large dataset for understanding urban scenes, featuring semantic, instance-wise, and pixel-level annotations across 30 classes grouped into 8 categories. It comprises around 5,000 finely annotated images and 20,000 coarsely annotated ones, recorded in various cities under different conditions. This dataset is valuable for tasks related to urban scene analysis.\\
\myparagraph{SUN RGB-D.}
The SUN RGB-D dataset~\cite{song2015sun} comprises 10,335 RGB-D images of room scenes, each with depth and segmentation maps. It's annotated for 700 object categories and divided into training and testing sets with 5,285 and 5,050 images, respectively. This dataset addresses the need for large-scale 3D annotations and metrics for scene understanding tasks. It includes data from four sensors, with extensive annotations for 2D and 3D object boundaries, orientations, room layout, and scene categories, enabling advanced algorithm training and cross-sensor bias study.\\
\myparagraph{nocaps.}
The nocaps~\cite{agrawal2019nocaps} dataset pushes image captioning models to grasp a wider array of visual concepts from diverse data origins. Comprising 166,100 human-generated captions for 15,100 images sourced from OpenImages, the dataset integrates different training data, including COCO image-caption pairs and OpenImages labels and bounding boxes, with a specific emphasis on describing objects.\\
\myparagraph{DRIVE.}
The DRIVE~\cite{staal2004ridge} dataset used for retinal vessel segmentation consists of 40 color fundus images, including 7 displaying abnormal pathology. Captured during diabetic retinopathy screenings in the Netherlands, these images were taken with a Canon CR5 camera featuring a 45-degree field of view. The dataset is split into a training set (20 images) and a testing set (20 images), each accompanied by a circular field of view (FOV) mask. Expert manual segmentations are provided for assessment in the training set, while the testing set includes two observer-based segmentations, with the first observer's results considered as the ground truth for evaluation.\\
\myparagraph{LoveDA.}
The LoveDA~\cite{wang2021loveda} dataset comprises 5987 high-resolution remote sensing images (0.3 m) from urban and rural areas in Nanjing, Changzhou, and Wuhan. It targets semantic segmentation and domain adaptation tasks, offering challenges such as multi-scale objects, complex backgrounds, and inconsistent class distributions, aiming to address diverse geographical environments.\\
\myparagraph{ISPRS Potsdam.}
The ISPRS Potsdam~\cite{potsdam} dataset comprises 38 patches with true orthophotos (TOP) and digital surface models (DSM) having a 5 cm ground sampling distance. The TOP images are available in various channel compositions (IRRG, RGB, RGBIR), and DSM files contain 32-bit float values representing heights. Some patches have normalized DSMs, indicating heights above the terrain. Ground truth labels are provided for a portion of the data, with the rest reserved for benchmark testing.\\
\myparagraph{WIDER Face.}
The WIDER Face~\cite{yang2016wider} dataset is a comprehensive face detection benchmark dataset, consisting of 32,203 images with a diverse range of 393,703 labeled faces. These images exhibit variations in scale, pose, and occlusion. The dataset is categorized into 61 event classes, with 40\% for training, 10\% for validation, and 50\% for testing. Evaluation follows the PASCAL VOC dataset metric.\\
\myparagraph{DeepFashion.}
The DeepFashion~\cite{liu2016deepfashion} dataset is a comprehensive collection of around 800,000 fashion images, accompanied by extensive annotations. These annotations include 46 fashion categories, 1,000 descriptive attributes, bounding boxes, and landmark information. The dataset covers a broad spectrum of fashion images, from well-posed product photos to real-world consumer snapshots.

\section{Training}\label{sec:train}
\subsection{Implementation Details}
\myparagraph{Training schemes.}
For single-task training, \ours-B$_{\text{single-task}}$ is typically trained using a batch size of 24 for 120,000 iterations on 8 NVIDIA A100 GPUs~(40GB), following a cosine annealing schedule. In multi-task joint training on five datasets, \ours-B$_{\text{multi-task}}$ undergoes training with the same batch size and GPU number for more iterations~(\textit{i.e.}, 640,000). 
The large and huge model variants require more GPU memory for training and are therefore trained on 12 and 24 GPUs, respectively. For large-scale universal training, we train all models using a batch size of 96 across 320,000 iterations. This process is conducted on setups of 32, 48, and 96 GPUs, resulting in total training times of 3, 5, and 7 days, respectively.\\
\myparagraph{Custom learning rate.}
For the layers without pretraining, we applied the standard base learning rate. In contrast, the layers that had been pretrained used progressively increasing learning rates. This strategy begins with a learning rate that is 0.1 times the base rate for the first pretrained layer, gradually escalating to a full 1.0 times the base rate by the final pretrained layer. We argue this method enhances the integration of pretrained and newly trained weights, leading to better overall performance of the model.\\
\myparagraph{Grid generation and sampling.}
We adjust the grid sizes according to the level of detail required by each task. For object detection and instance segmentation, we work with 5 $\times$ 5 grids in each window, while for semantic segmentation, we increase the grid size to 14 $\times$ 14. To illustrate, in object detection, an input image of 1120 $\times$ 1120 pixels is represented by a 25 $\times$ 25 grids, and in semantic segmentation, a 672 $\times$ 672 pixels is represented by a 42 $\times$ 42 grids. Computing losses for every point on these grids would demand excessive computational resources, particularly for semantic segmentation. To manage this, we employ a strategy of sampling specific grid points during training, selecting a predetermined number of points with a focus on including positive samples and supplementing with negative samples as needed. Specifically, for object detection and instance segmentation, we choose 10 points out of 25 in each window, and for semantic segmentation, we select 32 points out of 196.  As shown in Table~\ref{tab:grid_sampling}, this method effectively reduces computational costs without significant performance drops.
\vspace{-10pt}
\begin{table}[t]
  \centering
  \caption{Performance of grid sampling on object detection with 25 $\times$ 25 grid resolution.}
  \label{tab:grid_sampling}
  \resizebox{0.45\linewidth}{!}{
  \begin{tabular}{c|c|c}
    \toprule
    Sample Number & mAP & Training Time \\
    \midrule
    625 & 45.3& 47h \\
    250 & 45.1 & 20h \\
    \bottomrule
  \end{tabular}}
  \vspace{-20pt}
\end{table}
\subsection{Label Assignment}
\myparagraph{Object Detection.}
Our approach employs the well-established Hungarian matching algorithm~\cite{kuhn1955hungarian} for label assignment calculation. For each grid point, we compute its normalized L1 distance to the centers of all boxes as the matching cost.\\
\myparagraph{Instance Segmentation.}
Similar to object detection, instance segmentation targets are determined by computing the L1 distance between bounding box centers and grid positions. Polar coordinates with 24 rays, inspired by PolarMask~\cite{xie2020polarmask}, are employed for mask representation. The mass center of an object is calculated using its annotated polygon boundaries. Grid points classified as positive must accurately predict object category, bounding box, centroid, and distances from the mass center to boundary points. \\
\myparagraph{Semantic Segmentation.}
Expanding upon ViT, we generate patch features (42 $\times$ 42) by downsampling the image (672 $\times$ 672) via a factor of 16. Given the dense prediction nature of semantic segmentation, we align the grid point size with the patch feature size. To alleviate computational load, we downsample original mask annotations (672 $\times$ 672) by a factor of 4, resulting in annotations of size 168 $\times$ 168, which is four times larger than the grid size. Subsequently, each grid point autonomously predicts segmentation annotations for 16 positions within a 4 $\times$ 4 square centered around it.\\
\myparagraph{Image Captioning.}
In our image captioning process, we tokenize each caption into a fixed-length sequence of 20 tokens. If the caption length is shorter than 20 tokens, we pad it with termination symbols to ensure uniformity. \\
\myparagraph{Visual Grounding.}
In visual grounding tasks, each query directly targets a specific bounding box, removing the necessity to align boxes with grid points.
\begin{table*}[t]
  \centering
  \caption{The evaluation results of the models after universal training on five standard vision-centric benchmarks. 
  }
  \label{tab:universal_benchmark}
  \resizebox{0.8\textwidth}{!}{
  \begin{tabular}{l|c|ccc|ccc|c|cc|c}
    \toprule
    & &  \multicolumn{3}{c|}{Object Detection} & \multicolumn{3}{c|}{Instance Seg} & \multicolumn{1}{c|}{Semantic Seg} & \multicolumn{2}{c|}{Captioning} & \multicolumn{1}{c}{Grounding} \\
    \multirow{-2}{*}{Methods} & \multirow{-2}{*}{\#Params}  & AP & AP$_{50}$ & AP$_{75}$  & AP & AP$_{50}$ & AP$_{75}$  & mIoU(SS) & BLEU-4 & CIDEr & Acc@0.5 \\
    \midrule
    \ours-B$_{\text{universal}}$     & 131M & 44.4 & 61.2 &48.1 & 30.3 &53.0 & 30.0 & 44.6 & 33.6 & 108.3 & 84.2\\
    \ours-L$_{\text{universal}}$   & 387M & 50.2 & 67.6 &54.6 & 33.1 &58.4 & 32.7 & 48.1 & 36.2 & 117.5  & 86.0\\
    \ours-H$_{\text{universal}}$  & 756M & 53.3 & 71.2 &58.3 & 35.9 &62.6 & 36.1 & 53.0 & 37.7 & 124.2 & 88.3\\
    \toprule
  \end{tabular}
  }
\end{table*}
\subsection{Data Augmentation}
\myparagraph{Object Detection and Instance Segmentation.}
For object-level perception tasks, images undergo preprocessing steps. Initially, images are horizontally flipped with a 0.5 probability. Subsequently, two methods are employed to achieve a fixed input size. The first method involves direct resizing of the image to dimensions of 1120 $\times$ 1120, disregarding the original aspect ratio. The second method randomly resizes the image to one of three size pairs: (400, 4200), (500, 4200), or (600, 4200), while preserving the original aspect ratio. Following resizing, the image is cropped to a size of (384, 600) and then resized again to 1120 $\times$ 1120 pixels.\\
\myparagraph{Semantic Segmentation.}
In semantic segmentation, specific preprocessing steps are applied to images to ensure their size is standardized and to increase diversity. Initially, images are acquired with a size of 672 $\times$ 672 pixels, employing random selection between two methods. The first method directly resizes the image to 672 $\times$ 672, disregarding the original aspect ratio. The second method involves scaling the image to sizes ranging from 100\% to 200\% of 672, again without preserving the original aspect ratio. Following this, a random crop is applied to ensure the image size remains 672 $\times$ 672 pixels. Moreover, to augment image diversity, two additional operations are performed with a 50\% probability: horizontal flipping and photometric distortions. These steps collectively contribute to a more robust dataset for segmentation tasks.\\
\myparagraph{Image Captioning.}
As for this task, we initiate preprocessing with a dynamic crop, varying size ratio in [0.08, 1.0] and aspect ratio in [3/4, 4/3] in relation to the original image. Following this crop, the image is resized to 224$\times$224 dimensions. Additionally, there is a 50\% probability of horizontally flipping the image for further augmentation.\\
\myparagraph{Visual Grounding.}
Visual grounding augmentation includes color adjustments with a 50\% probability, enabling changes in brightness, contrast, saturation, and hue. Subsequently, the image undergoes a random crop within a relative range of (0.8, 0.8) of the original size. Finally, we resize the image to 224$\times$224 without keeping the original aspect ratio.
\begin{table*}[t]
  \centering
  \caption{Universal training evaluation results on detection, instance segmentation, and visual grounding datasets.
  }
  \label{tab:universal_3tasks}
  \resizebox{\textwidth}{!}{
  \begin{tabular}{l|ccccc|cccc|c}
    \toprule
    &    \multicolumn{5}{c|}{Object Detection@AP} &    \multicolumn{4}{c|}{Grounding@Acc} & Instance Seg@AP \\
    \multirow{-2}{*}{Methods}   & Objects365~\cite{shao2019objects365}& OpenImages~\cite{kuznetsova2020open} & LVIS~\cite{gupta2019lvis}  & VOC0712~\cite{everingham2010pascal} & nuImages~\cite{caesar2020nuscenes} & 
    RefCOCO+~\cite{kazemzadeh2014referitgame} & RefCOCOg~\cite{mao2016generation} & Flickr30K~\cite{plummer2015flickr30k} & RefCLEF~\cite{kazemzadeh2014referitgame} &
    LVIS~\cite{gupta2019lvis} \\
    \midrule
    \ours-B$_{\text{universal}}$      & 17.7 & 43.4 &12.3 & 79.0 & 44.5 & 72.5 &76.9  & 71.0 & 72.2 &8.4 \\
    \ours-L$_{\text{universal}}$  & 25.5 & 51.6 &17.3 & 83.6 &47.2  & 73.9  & 78.9 &72.7 &74.5 & 11.4\\
    \ours-H$_{\text{universal}}$   & 31.9 & 57.7 &21.7 & 84.9 &50.0   & 78.3 & 80.7&77.5 &75.8 & 14.8 \\
    \toprule
  \end{tabular}
  }
\end{table*}
\begin{table*}[t]
  \centering
   \caption{Evaluation of universal training on segmentation datasets, with all results measured using the mIoU metric.
  }
  \label{tab:universal_segmentation}
  \resizebox{0.8\linewidth}{!}{
  \begin{tabular}{l|cccc}
    \toprule
    Methods   & COCO-Stuff~\cite{caesar2018coco}  & Pascal Context~\cite{mottaghi2014role} & BDD100K~\cite{yu2020bdd100k} & Mapillary Vistas~\cite{neuhold2017mapillary}   \\
    \midrule
    \ours-B$_{\text{universal}}$     & 42.6  &56.8 & 57.8 &23.0\\
    \ours-L$_{\text{universal}}$   & 46.0 &60.4 & 59.3 &25.4 \\
    \ours-H$_{\text{universal}}$   & 49.1 &63.3 & 61.5 &28.9 \\
    \toprule
  \end{tabular}
  }
  \vspace{-10pt}
\end{table*}
\begin{table}[t]
  \centering
  \caption{Decoding steps for all five tasks.}
  \label{tab:decode_step}
  \resizebox{0.95\linewidth}{!}{
  \begin{tabular}{c|c|c|c|c|c}
    \toprule
    Task & Object Detection & Instance Segmentation & Semantic Segmentation & Image Captioning & Visual Grounding \\
    \midrule
    Decoding Step & 5 & 31 & 16 &20&4 \\
    \bottomrule
  \end{tabular}
  }
  \vspace{-2pt}
\end{table}
\section{Evaluation}\label{sec:inference}
\subsection{Auto-regressive Decoding}
We tailor unique decoding rules for various tasks based on task templates. For example, in object detection, using the template { \texttt{$<$c$>$}\texttt{$<$$\text{x}_1$$>$}\texttt{$<$$\text{y}_1$$>$}\texttt{$<$$\text{x}_2$$>$}\texttt{$<$$\text{y}_2$$>$}, the category is decoded in the first position, drawing from a vocabulary containing all categories in the dataset. The subsequent four positions decode numerical values, drawing from a vocabulary of discretized locations. Table~\ref{tab:decode_step} illustrates the fixed decoding step number for all tasks, with no terminator token required except for image captioning. In image captioning, predictions following the terminator are disregarded during inference. 
\subsection{Inference Speed}\label{sec:speed}
In Table~\ref{tab:compute}, we present the inference speed of \ours-B across five tasks, measured on a single NVIDIA A100 GPU with a batch size of 1. Due to our adherence to the auto-regressive decoding paradigm commonly seen in NLP, we inherit the drawback of slow inference speed. This limitation becomes more pronounced in high-resolution object-level and semantic segmentation tasks that necessitate per-pixel predictions. However, we contend that leveraging multiple parallel decoding has significantly improved our method's speed, bringing it to an acceptable level. As shown in Table~\ref{tab:speed_compare}, our approach demonstrates comparable segmentation speed to SAM. Given that our structure and prediction approach closely align with LLM, the inference acceleration techniques~\cite{chen2023accelerating} employed for LLM also hold promise for enhancing our method.
\begin{table}[t]
    \centering
    \begin{minipage}{0.56\linewidth}
        \centering
        \caption{Inference speed of \ours-B on A100.}
        \label{tab:compute}
        \resizebox{0.9\textwidth}{!}{
      \begin{tabular}{c|c|c|c|c}
    \toprule
    Task & Resolution& Grid Number & Decoding Step& FPS\\
    \midrule
    Object Detection & 1120 $\times$ 1120& 625 & 5 & 2.5 \\
    Instance Segmentation & 1120 $\times$ 1120 &625 &31 &0.7\\
    Semantic Segmentation & 672 $\times$ 672 & 1764 & 16 &1.5 \\
    Image Captioning & 224 $\times$ 224 & 1 & 20 & 3.2\\
    Visual Grounding & 224 $\times$ 224 & 1 & 4&8.1\\
    
    \toprule
  \end{tabular}
  }
    \end{minipage}
    \hfill
    \begin{minipage}{0.43\linewidth}
      \centering
      \caption{Latency comparison with SAM on semantic segmentation task.}
      \label{tab:speed_compare}
      \resizebox{\textwidth}{!}{
      \begin{tabular}{c|c|c|c}
            \toprule
            Method~(ADE20K) & Resolution & \#Params & FPS \\
            \midrule
            SAM-B~[{\color{green}41}] & 672 $\times$ 672 & 90M& 1.6 \\
            \ours-B & 672 $\times$ 672 & 131M & 1.5  \\
            \bottomrule
            \end{tabular}}
            \vspace{2pt}
    \end{minipage}
    \vspace{-16pt}
\end{table}
\subsection{Benchmarking Setup} \label{sec:evaluation}
\textbf{Multi-Task Learning.} On the multi-task datasets, we conducted evaluations on the validation sets, except for COCO Caption~\cite{caesar2018coco}, where we used the Karpathy split~\cite{karpathy2015deep} for evaluation on the test set.\\
\textbf{Universal Learning.} We evaluate our universal models on several key datasets. Table \ref{tab:universal_benchmark} presents their performance on representative datasets for five tasks. However, due to the less frequent sampling of these analyzable multi-task datasets during universal training, their performance slightly lags behind models trained on multi-task benchmark. For further performance insights on other datasets, refer to Tables \ref{tab:universal_3tasks} and \ref{tab:universal_segmentation}. Notably, for image captioning, all datasets except COCO Caption are entirely used in training, obviating the need for extra evaluation.\\
\textbf{Few-shot Learning.} We adopt the classical N-way K-shot~\cite{finn2017model} setting to create a support set for few-shot evaluation. In this setup, for each class in the dataset, we extract k samples labeled with the corresponding class, resulting in the selection of N$\times$K samples. By default, K is set to 5.  As depicted in Table \ref{tab:few_shot_datasets}, we sample varying quantities of support sets depending on the number of categories in each dataset. Each experiment, by default, iterates 100 times on the support set. However, due to the limited size of the support set in WIDERFace~\cite{yang2016wider}, we reduce the iteration count to 50 times to mitigate the risk of overfitting. All few-shot training is conducted with a fixed learning rate of 2e-4.\\
\indent We select Faster R-CNN~\cite{ren2015faster} and DeepLabV3~\cite{chen2018deeplabv3+}, two classic methods, as comparative baselines. In the case of Faster R-CNN, we employ the version with ResNet-50 as the backbone, utilizing pre-trained weights from the COCO~\cite{lin2014coco} dataset. For DeepLabV3, we opt for the version with ResNet-101 as the backbone, leveraging pre-training on the ADE20K~\cite{zhou2017scene} dataset.
\begin{table}[t]
  \caption{Few shot datasets. }
  \label{tab:few_shot_datasets}
  \centering
  \resizebox{0.8\linewidth}{!}{
  \begin{tabular}{l|c|c|c|c}
    \toprule
    Dataset  & Size & Category Number & Support Set Size & Training Iters  \\
    \midrule
    DRIVE~\cite{staal2004ridge}  & 40 & 2 & 10 & 100 \\
    LoveDA~\cite{wang2021loveda} & 5,987& 7& 35 & 100 \\
    ISPRS Potsdam~\cite{potsdam} &5,472 &6& 30 & 100\\
    WIDERFace~\cite{yang2016wider} & 32,203&1&5 & 50\\
    DeepFashion~\cite{liu2016deepfashion} &800,000 & 15 & 75 & 100\\
    \bottomrule 
  \end{tabular}
  }
\vspace{-4pt}
\end{table}
\begin{table}[t]
    \centering
    \begin{minipage}{0.50\linewidth}
        \centering
        \caption{Ablation of text conditioning on visual grounding task.}
        \label{tab:ablate_text_conditioning}
        \resizebox{0.8\textwidth}{!}{
      \begin{tabular}{c|c|c}
        \toprule
        Models &Text Conditioning &  Acc@0.5 \\
        \midrule
           \ours-B$_{\text{single-task}}$ &  &82.7  \\
         \ours-B$_{\text{single-task}}$ &\checkmark & 83.3\\
        \ours-B$_{\text{multi-task}}$&  &78.6 \\
        \ours-B$_{\text{multi-task}}$ & \checkmark &85.8 \\
        \toprule
      \end{tabular}
      }
    \end{minipage}
    \hfill
    \begin{minipage}{0.47\linewidth}
      \centering
      \caption{Ablation of beam number on image captioning task.}
      \label{tab:ablate_beam}
      \resizebox{0.6\textwidth}{!}{
      \begin{tabular}{c|c|c}
        \toprule
        Beam Number &BLEU-4 &  CIDEr \\
        \midrule
        1 & 33.1 & 106.9  \\
        2  &33.5 & 107.2 \\
        3 & 33.7  & 107.9 \\
        5 & 33.7 & 107.6 \\
        \toprule
      \end{tabular}
      }
    \end{minipage}
    \vspace{-16pt}
\end{table}
\section{More ablation studies}\label{sec:ablate}
\myparagraph{Text Conditioning.}
In visual grounding, we incorporate image-to-text attention during network forwarding, enhancing task differentiation between detection and visual grounding. Table \ref{tab:ablate_text_conditioning} demonstrates that incorporating text conditioning results in a modest improvement of +0.6 in visual grounding when trained independently. However, its impact becomes more significant in multi-task training, showing a remarkable enhancement of +7.2, aligning with our hypothesis.\\
\myparagraph{Beam Search.} Table~\ref{tab:ablate_beam} demonstrates how performance varies with different beam sizes in beam search. We observe an improvement as the beam size increases from 1 to 2, but performance stabilizes between 2 and 5, with only a minor drop in CIDEr. Given that larger beam sizes lead to longer inference times, we have selected a default beam size of 2. \\
\myparagraph{Mass Center and Ray Number.} Table~\ref{tab:ablate_ray} presents an ablation of instance segmentation settings. Utilizing the mass center yields better results than the box center, probably because the box center might fall outside the object. Employing 36 rays slightly improves performance but at the cost of significant training time.

\section{Specific Modules of Comparison Methods}\label{sec:specifc}
In Table \ref{tab:specific}, we outline the specific modules and parameter quantities utilized for method comparison. Many methods, regardless of whether they are specialist or generalist models, incorporate task-specific modules and modality-specific encoders in their designs. In contrast, our approach is characterized by its simplicity, as it does not rely on such intricate designs.
\begin{table}[t]
  \centering
  \caption{Ablation on instance segmentation settings.}
  \label{tab:ablate_ray}
  \resizebox{0.5\textwidth}{!}{
  \begin{tabular}{c|c|c|c|c}
    \toprule
    Box Center & Mass Center & Ray Number & mAP & Training Time \\
    \midrule
     \checkmark&& 24 & 29.0 & 32h \\
     \checkmark& & 36  &29.2 & 49h  \\
    &\checkmark & 24& 31.4 & 32h   \\
    &\checkmark &36 &  31.7 & 49h  \\
    \toprule
  \end{tabular}
  }
\end{table}
\begin{table*}[t]
  \centering
  \caption{Specific modules and their corresponding parameter quantities for the methods used for comparison. The parameter of text embedding is excluded because it operates in a zero-computation index manner.}
  \label{tab:specific}
  \resizebox{0.95\linewidth}{!}{
  \begin{tabular}{l|c|c|c|}
    \toprule
    Methods &Specific Modules & Num &  \#Params  \\
    \midrule
    \multicolumn{4}{l}{\textbf{\textit{Specialist Models}}} \\
    Faster R-CNN-FPN~\cite{ren2015faster} & ResNet,FPN,RPN,ClassficationHead,RegressionHead & 5   & 42M    \\
    DETR-DC5~\cite{carion2020detr}        & ResNet,Encoder,Decoder,ClassficationHead,RegressionHead & 5   & 41M       \\
    Deformable-DETR~\cite{zhu2020deformabledetr}  & ResNet,Encoder,Decoder,ClassficationHead,RegressionHead & 5  & 40M   \\
    Mask R-CNN~\cite{he2017mask}       & ResNet,FPN,RPN,RPNHead,ClassficationHead,RegressionHead & 6  & 46M     \\
    Polar Mask~\cite{xie2020polarmask} & ResNet,FPN,ClassficationHead,CenternessHead,RegressionHead & 5  & 55M     \\
    Mask2Former~\cite{cheng2021mask2former}  & ResNet,PixelDecoder,TransformerDecoder,ClassficationHead,MaskHead & 5    & 44M      \\
    Pix2Seq~\cite{chen2021pix2seq}        & ResNet,Encoder,Decoder & 3   & 37M   \\
    UNITER ~\cite{chen2020uniter}         & Faster~R-CNN,Project Layer, Encoder,Decoder & 4 & 303M     \\
    VILLA   ~\cite{gan2020large}         & Faster~R-CNN, Encoder,Decoder & 3  & 369M   \\
    MDETR  ~\cite{kamath2021mdetr}       &CNN,RoBERTa,Image Adapter, Text Adapter,Encoder,Decoder& 6 & 188M    \\
    VL-T5 ~\cite{cho2021unifying}        & Faster R-CNN, Encoder,Decoder & 3  & 440M  \\
    DeepLabV3+~\cite{chen2018deeplabv3+}       &ResNet,Decoder,Auxiliary Head& 3 & 63M \\
    TokenFusion~\cite{wang2022tokenfusion}   &Segformer,YOLOS,Fusion Module,GroupFree & 4  & 79M \\
    U-Net~\cite{ronneberger2015unet} & Encoder,Decoder,Decode Head& 3 &8M\\
    AerialFormer~\cite{yamazaki2023aerialformer} & Transformer Encoder, CNNs Stem,  Multi-Dilated CNNs Decoder &3 &114M \\
    RetinaFace~\cite{deng2020retinaface} &ResNet,FPN,ClassficationHead,RegressionHead,ContextModule & 5 & 30M\\
    \midrule
    \multicolumn{4}{l}{\textbf{\textit{Generalist Models}}} \\
    UniTab ~\cite{yang2022unitab}          & Image Encoder,Text Encoder, Multimodal Encoder, Decoder & 4  & 185M  \\
    Uni-Perceiver ~\cite{zhu2022uni}       & None & 1  & 124M      \\
    Uni-Perceiver-MoE  ~\cite{zhu2022uni-moe}   & None & 1  & 167M   \\
    Uni-Perceiver-V2 ~\cite{li2023uni}     & ResNet,RPN,Mask~DINO,RoBERTa,Decoder,ClassficationHead,RegressionHead,MaskHead& 8 & 308M  \\
    Pix2Seq v2~\cite{chen2022unified}      & ViT,Decoder & 2  & 132M   \\
    Unified-IO$_{XL}$~\cite{lu2023unifiedio} & VQ-VAE Encoder,VQ-VAE Decoder,Encoder,Decoder & 4  & 2.9B   \\
    Shikra-13B~\cite{chen2023shikra} & ViT,Vicuna,Image Adapter & 3 & 13B   \\
    Ferret-13B~\cite{you2023ferret} & ViT,Vicuna,Visual Sampler,KNN & 4 & 13B   \\
    VisionLLM-R50 ~\cite{wang2023visionllm}   &ResNet,Language-Guided Image Tokenizer,Encoder,Decoder,Alpaca-7B& 5  & 7B \\
    
    GLIP-T~\cite{li2022glip}  &Swin,FPN,Text Encoder,Dy-Head,Fusion Module &5  &431M \\
    Grounding DINO-T~\cite{liu2023groundingdino}  &Swin,DINO,BERT,Feature Enhancer,Decoder,Query Selection &6 &174M  \\
    BLIP~(129M) ~\cite{li2022blip}  &ViT-L,BERT,Image-grounded Text Encoder, Image-grounded Text Decoder& 4& 583M  \\
    BLIP-2~(129M) ~\cite{li2023blip}  &ViT-G,Qformer,Adapter,LLM& 4 & 12.1B\\
    ReCo+~\cite{shin2022reco} &DeiT-SIN,CLIP,DenseCLIP,DeepLabV3+ & 4 &46M \\
    XDecoder(T)~\cite{zou2023xdecoder}  &FocalNet,Encoder,Decoder,Latent Query & 4 &165M  \\
    \toprule
  \end{tabular}
  }
\end{table*}
\section{Visualization} \label{sec:vis}
\myparagraph{Task Visualization.} In Figure \ref{fig_task_vis}, we visualize an example for each task, showcasing the image input, text-formatted predictions, and the visualization of the prediction results from left to right. 
For simplicity, we selected a few examples of local responses predicted by the model and listed their corresponding text-formatted predictions.\\
\myparagraph{Zero-shot Visualization.} In Figure \ref{fig_zero_vis}, we showcase qualitative examples of predictions on zero-shot datasets made by \ours-H$_{\text{universal}}$. Notably, our model accurately predicts missing annotations in some cases. For instance, in Cityscapes detection, it correctly identifies unannotated bicycles and vehicles, even under low-light conditions. A similar accuracy is observed in SUN RGB-D segmentation, where the model detects all chairs, although only two are annotated. In Cityscapes segmentation, despite the dataset's bias of excluding self-owned vehicles from annotation, our model demonstrates exceptional generalization by correctly classifying these vehicles, relying on minimal information and without dataset-specific fine-tuning.\\
\myparagraph{Few-shot Visualization.} Figure \ref{fig_few_vis} provides visual representations of the qualitative predictions made by \ours-H$_{\text{universal}}$ on few-shot datasets. These examples highlight the remarkable performance of our model in situations with limited data, emphasizing its potential for applications across diverse domains.
\begin{figure*}[t]
    \begin{center}
    \includegraphics[width=\textwidth]{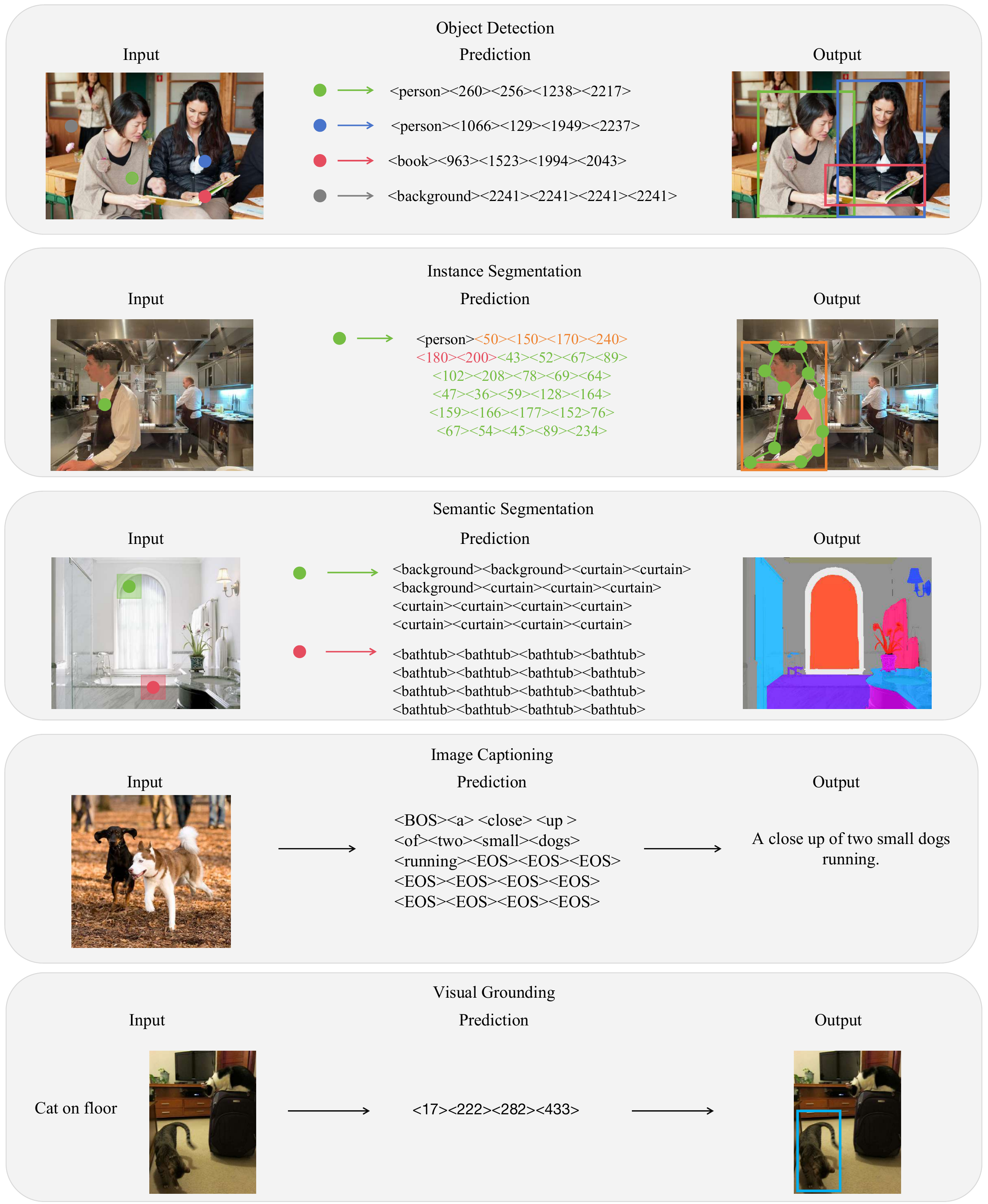}
    \end{center}
    \caption{Visualization of five standard vision-centric tasks.}
    \vspace{-12pt}
    \label{fig_task_vis}
\end{figure*}
\begin{figure*}[t]
    \begin{center}
    \includegraphics[width=\textwidth]{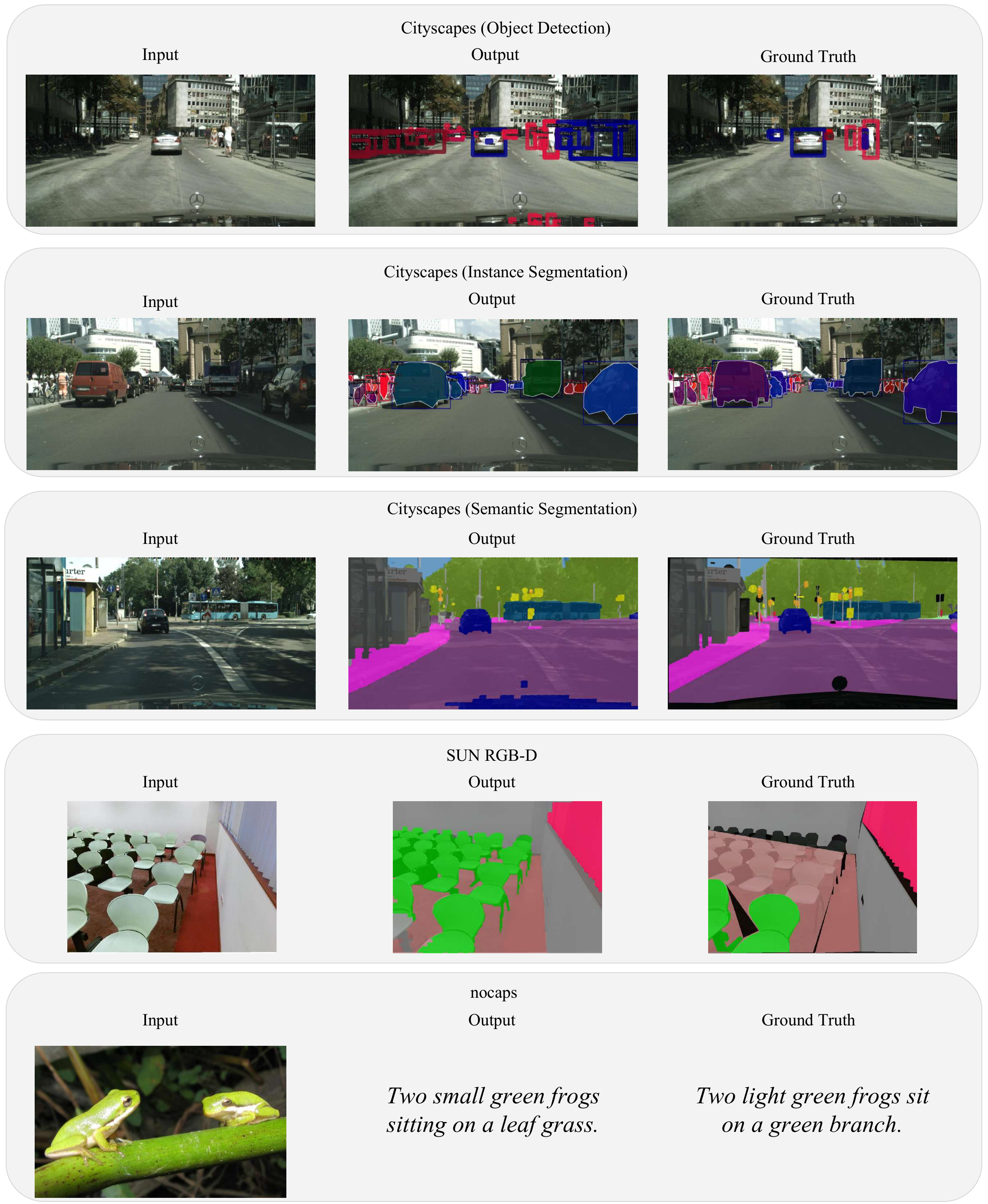}
    \end{center}
    \vspace{-12pt}
    \caption{Qualitative results on zero-shot datasets. Zoom in for better viewing.}
    \vspace{-12pt}
    \label{fig_zero_vis}
\end{figure*}
\begin{figure*}[t]
    \begin{center}
    \includegraphics[width=\textwidth]{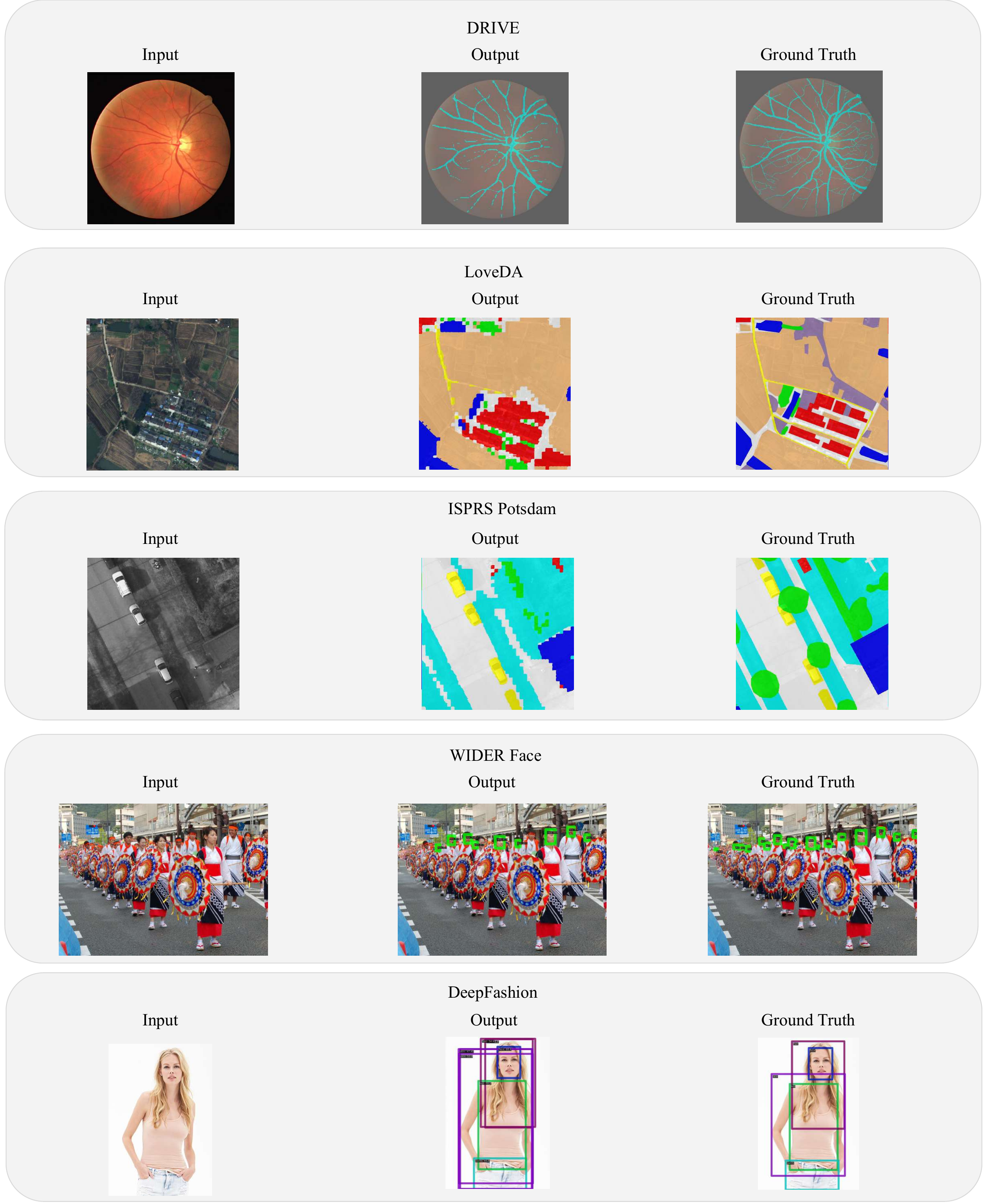}
\end{center}
    \vspace{-12pt}
    \caption{Qualitative results on few-shot datasets. Zoom in for better viewing.}
    \vspace{-12pt}
    \label{fig_few_vis}
\end{figure*}
\section{Discussion} \label{sec:discussion}
\myparagraph{Comparison with Fuyu-8B.} Compared to Fuyu-8B~\cite{fuyu-8b}, which focuses on well-explored vision-language
tasks, our \ours extends the scope of the multi-layer transformer to often-overlooked object and pixel-level tasks with a universal language interface. To achieve it, we design a flexible parallel decoding template using point prompts for task unification across various perceptual scales. The local image prompt is also introduced to enhance fine-grained perception ability. \\
\myparagraph{Comparison with adapter-based methods.}  Our method provides an
alternative solution for LVMs. Unlike previous fine-tuning efforts with LLMs, we aim to close the architectural gap between vision and language. Moreover, our \ours allows easy end-to-end implementation without module-specific design, greatly simplifying the training process and model scaling. \\
\myparagraph{Limitations.} Constrained by training data limited to five selected tasks with relatively straightforward task prompts, GiT struggles to generalize to entirely new tasks in zero-shot settings. Task-level zero-shot remains challenging, even for capable LLMs. \ours closely aligns with it and inherits this limitation. However, our GiT shows strong extendibility in task unification, potentially supporting various other tasks by incorporating relevant data. \\
\myparagraph{Negative Societal Impact.} Our largest model necessitates 7 days of training on 96 A100 GPUs, leading to considerable carbon emissions. Furthermore, the generated content might reflect biases from the training data, stemming from a lack of alignment with human preferences.

\end{document}